\documentclass[10pt,twocolumn,letterpaper]{article}

\usepackage{cvpr}              %

\usepackage[dvipsnames]{xcolor}

\definecolor{cvprblue}{rgb}{0.21,0.49,0.74}
\usepackage[pagebackref,breaklinks,colorlinks,citecolor=cvprblue]{hyperref}

\usepackage{adjustbox}
\usepackage{multirow}
\usepackage{colortbl}
\usepackage{hyperref}
\usepackage{cuted}

\newcommand{\mypar}[1]{\vspace{-3mm}\paragraph{#1}}
\newcommand{\x}[0]{{\mathbf x}}
\newcommand{\A}[0]{{\mathbf A}}
\def\paperID{6108} %
\def\confName{CVPR}
\def\confYear{2024}

\title{Visual Anagrams: Generating Multi-View \\Optical Illusions with Diffusion Models}

\author{Daniel Geng \qquad Inbum Park \qquad Andrew Owens\\
University of Michigan\\    
{\url{https://dangeng.github.io/visual_anagrams/}}\vspace{-3em}
}

\begin{document}

\maketitle

\begin{strip}
\includegraphics[width=1\linewidth]{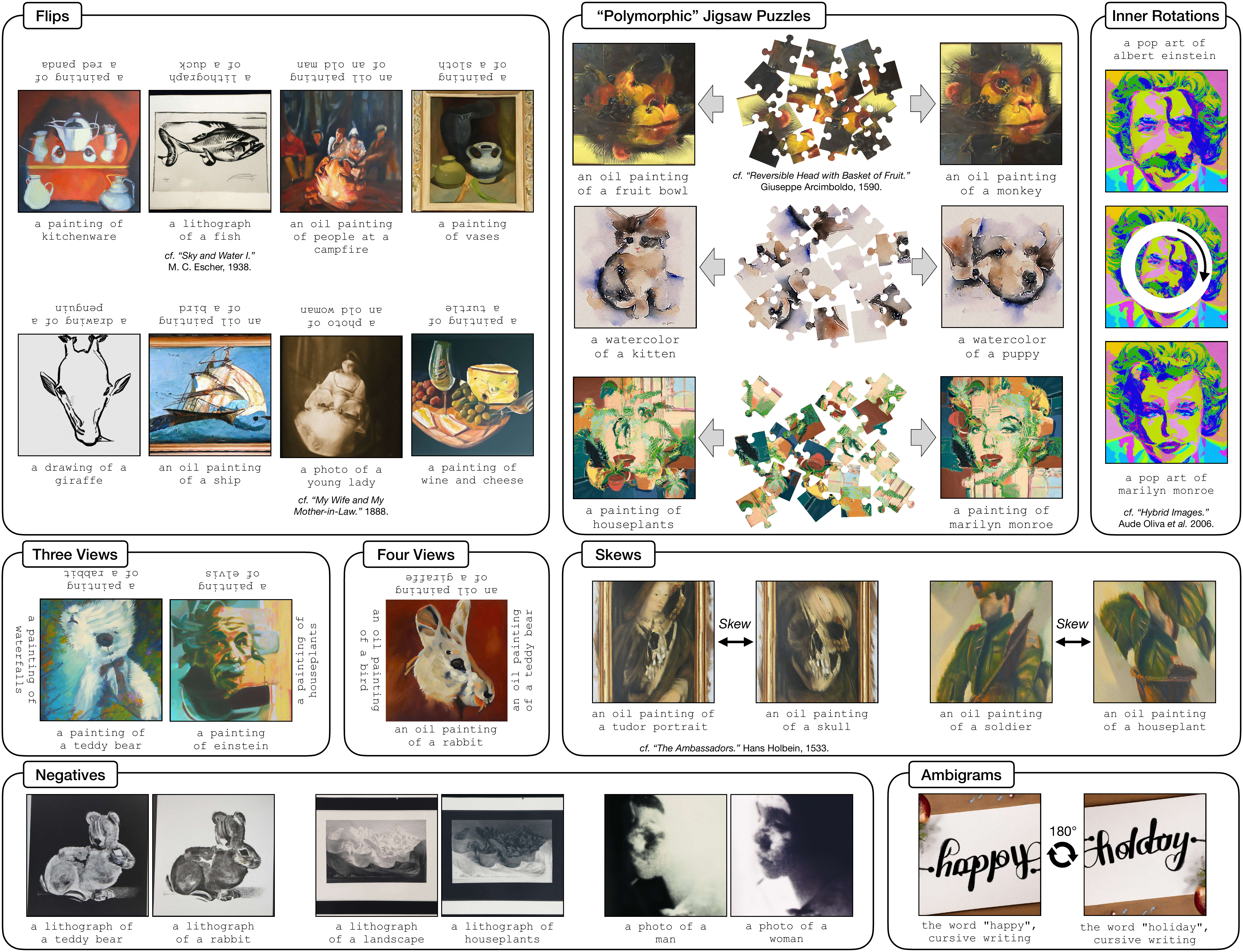} \vspace{-4mm}
\vspace{0.5mm}
\captionof{figure}{{\bf Generating Multi-View Illusions.} We propose a method for generating optical illusions from an off-the-shelf text-to-image diffusion model. We create images that match different prompts after undergoing a transformation. Our approach supports a variety of transformations, including flips, rotations, skews, color inversions, and jigsaw rearrangements. All images are hand selected. For random samples, please see \cref{fig:random} and Appendix~\ref{sec:apdx_random}. \textbf{For easier viewing, please see our \href{https://dangeng.github.io/visual_anagrams/}{webpage} for animated versions of these illusions.} %
\vspace{-1em}}

\label{fig:teaser} 
\end{strip}

\begin{abstract}

We address the problem of synthesizing multi-view optical illusions: images that change appearance upon a transformation, such as a flip or rotation. We propose a simple, zero-shot method for obtaining these illusions from off-the-shelf text-to-image diffusion models. 
During the reverse diffusion process, we estimate the noise from different views of a noisy image, and then combine these noise estimates together and denoise the image.
A theoretical analysis suggests that this method works precisely for views that can be written as orthogonal transformations, of which permutations are a subset. This leads to the idea of a {\em visual anagram}---an image that changes appearance under some rearrangement of pixels. This includes rotations and flips, but also more exotic pixel permutations such as a jigsaw rearrangement. 
Our approach also naturally extends to illusions with more than two views.
We provide both qualitative and quantitative results demonstrating the effectiveness and flexibility of our method. Please see our project webpage for additional visualizations and results: \url{https://dangeng.github.io/visual_anagrams/}

\end{abstract}
\vspace{-1em}

\section{Introduction}
\label{sec:intro}

Images that change their appearance under a transformation, such as a rotation or a flip, have long fascinated students of perception, from Salvador Dalí to M. C. Escher. The appeal of these {\em multi-view optical illusions} lies partly in the challenge of arranging visual elements such that they may be understood in multiple different ways. Creating these illusions requires accurately modeling---and then subverting---visual perception.

In this paper, we propose a simple, zero-shot method for creating multi-view illusions with off-the-shelf text-to-image diffusion models. 
In contrast to most previous work on computationally generating optical illusions~\cite{freeman1991motion,oliva2006hybrid,chu2010camouflage,chandra2022designing,gomez2022synthesis,hirsch2020color,chi2014optical,makowski2021parametric,shinbrot2017network,ehm2011variational,owens2014camouflaging,guo2023ganmouflage}, our method does not require an explicit model of human perception. Rather, our approach builds on work that suggests generative models may process optical illusions in a way similar to humans~\cite{jaini2023intriguing,ngo2023clip,gomez2019convolutional}. %
In this way, our method is similar to recent work that uses diffusion models to create optical illusions by Burgert~\etal~\cite{burgert2023illusions} and Tancik~\cite{tancik2023illusions}.

Our method can generate many types of classic illusions, such as images that change appearance when flipped or rotated (\cref{fig:teaser}), as well as a new class of illusions which we term {\em visual anagrams}. These are images that change appearance under a permutation of their pixels. Image flips and rotations are a subset of these, as they can both be expressed as a permutation of pixels, but we also consider more exotic permutations. For example, we generate jigsaw puzzles that can be solved in two different ways, which we call ``polymorphic jigsaws." In addition, we successfully apply our approach to generating illusions with three and four views (\cref{fig:teaser}).

Our method works by using a diffusion model to denoise an image from multiple views, obtaining multiple noise estimates. These noise estimates are then combined to form a single noise estimate which is used to perform a step in the reverse diffusion process. However, we show that care must be taken in choosing these views. For one, the transformation must keep the statistics of the noise intact, as the diffusion model is trained under the assumption of i.i.d.\ Gaussian noise. We provide an analysis of these conditions and give an exact specification of the class of transformations supported. Our contributions are as follows:

\begin{itemize}
    \item We present a simple yet effective method for generating multi-view optical illusions using diffusion models.
    \item We derive a precise description of the set of views that our method supports and provide empirical evidence that these views work.
    \item We consider practical design decisions, crucial to optimizing the quality of generated illusions, and report ablations on our choices.
    \item We provide quantitative and qualitative results, showcasing both the efficacy and flexibility of our method.
\end{itemize}

\section{Related Work}
\label{sec:related_work}

\paragraph{Diffusion Models.}  Diffusion models~\cite{sohldickstein2015diffusion,ho2020denoising,song2021scorebased,dhariwal2021diffusion,graikos2022diffusion,ramesh2021zero,rombach2022ldm,saharia2022imagen,song2020denoising} are a class of powerful generative models that iteratively convert a sample from a noise distribution to a sample from some data distribution. These models work by estimating the noise in a noisy sample, and removing the estimated noise following some update rule such as DDPM~\cite{ho2020denoising} or DDIM~\cite{song2020denoising}. A prominent application of diffusion models has been text-conditioned image synthesis~\cite{nichol2021glide,rombach2022ldm,deepfloyd2023,saharia2022imagen}. In addition to a noisy image and a timestep, these models take a language model embedding of a text prompt as conditioning. Our approach is closely related to recent works that experiment with composing energy-based models and diffusion models~\cite{du2020compositional,liu2021learning,liu2022compositional,du2023reduce,garipov2023compositional,du2019implicit}. These approaches~\cite{liu2022compositional,du2023reduce} have shown that noise estimates from multiple conditional distributions can be combined together to obtain samples from compositions of the learned distributions. Our method uses a similar approach, and we apply it to the problem of multi-view illusion generation.

\mypar{Computational Optical Illusions.} Optical illusions serve as a testbed for understanding both human and machine perception~\cite{hertzmann2020visual,wang2020toward,gomez2019convolutional,ngo2023clip,jaini2023intriguing}. We focus on generating illusions computationally, an area which has primarily relied on models of how our brains process external stimuli. Freeman~\etal~\cite{freeman1991motion} create the illusion of constant motion in a desired direction by locally applying a filter with continuously shifting phase, relying on the observation that local phase-shifts are interpreted as global movement. Oliva~\etal~\cite{oliva2006hybrid} propose a method to make ``hybrid images," which change appearance depending on the distance they are viewed from. This method takes advantage of the multiscale nature of human perception by blending high frequencies of one image with low frequencies from another. Chu~\etal~\cite{chu2010camouflage} camouflage objects in a scene through re-texturing, with additional constraints on luminance as to preserve salient features of the object, and other work camouflages objects from multiple viewpoints in 3D scenes~\cite{owens2014camouflaging,guo2023ganmouflage}. Recently, Chandra~\etal~\cite{chandra2022designing} design color-constancy, size constancy, and face perception illusions by differentiating through a Bayesian model of human vision. Our method likewise generates illusions, but does not depend on an explicit model of human perception. Instead, our method works by leveraging visual priors in diffusion models learned implicitly through data. This aligns with observations~\cite{jaini2023intriguing,ngo2023clip,gomez2019convolutional} that generative models process illusions similarly to humans, and predict the same ambiguities. From this perspective, we can view our method as leveraging generative, rather than discriminative, models to synthesize adversarial examples~\cite{goodfellow2014explaining} against humans~\cite{elsayed2018adversarial}.

\mypar{Illusions with Diffusion Models.}
Very recently, artists and researchers have taken steps that show the potential of using diffusion models to create illusions. An artist under the pseudonym MrUgleh~\cite{ugleh2023spiral} repurposed a model fine-tuned for generating QR codes~\cite{qrmonster2023,zhang2023controlnet} to create images whose global structure subtly matches a given template image. In contrast, we study multi-view illusions that can be created zero-shot from off-the-shelf diffusion models, and our illusions are specified via text rather than images. Burgert~\etal~\cite{burgert2023illusions} use score distillation sampling (SDS)~\cite{poole2022dreamfusion,wang2023score} to create images that align with different prompts from different views. While in principle this approach supports a superset of our views, the use of SDS results in significantly lower quality results, and the need for explicit optimization leads to long sampling times. 
Our method is most similar to a proof-of-concept by Tancik~\cite{tancik2023illusions}, which creates rotation illusions by sampling from a latent diffusion model~\cite{rombach2022ldm} while alternating noise estimates between different views and prompts. While our technical approach is similar, by contrast we systematically study multi-view illusions, both by experimentally evaluating many different types of illusions and by providing a theoretical analysis of which views are (and are not) supported. In doing so, we go beyond just rotation views. We also make a number of improvements that result in qualitatively and quantitatively better illusions, such as by identifying a source of artifacts from latent diffusion, and  by adding support for an arbitrary number of views. To our knowledge, we are the first to systematically evaluate illusions generated by these approaches.

\section{Method}

\begin{figure}[t!]
    \centering
    \includegraphics[width=1\linewidth]{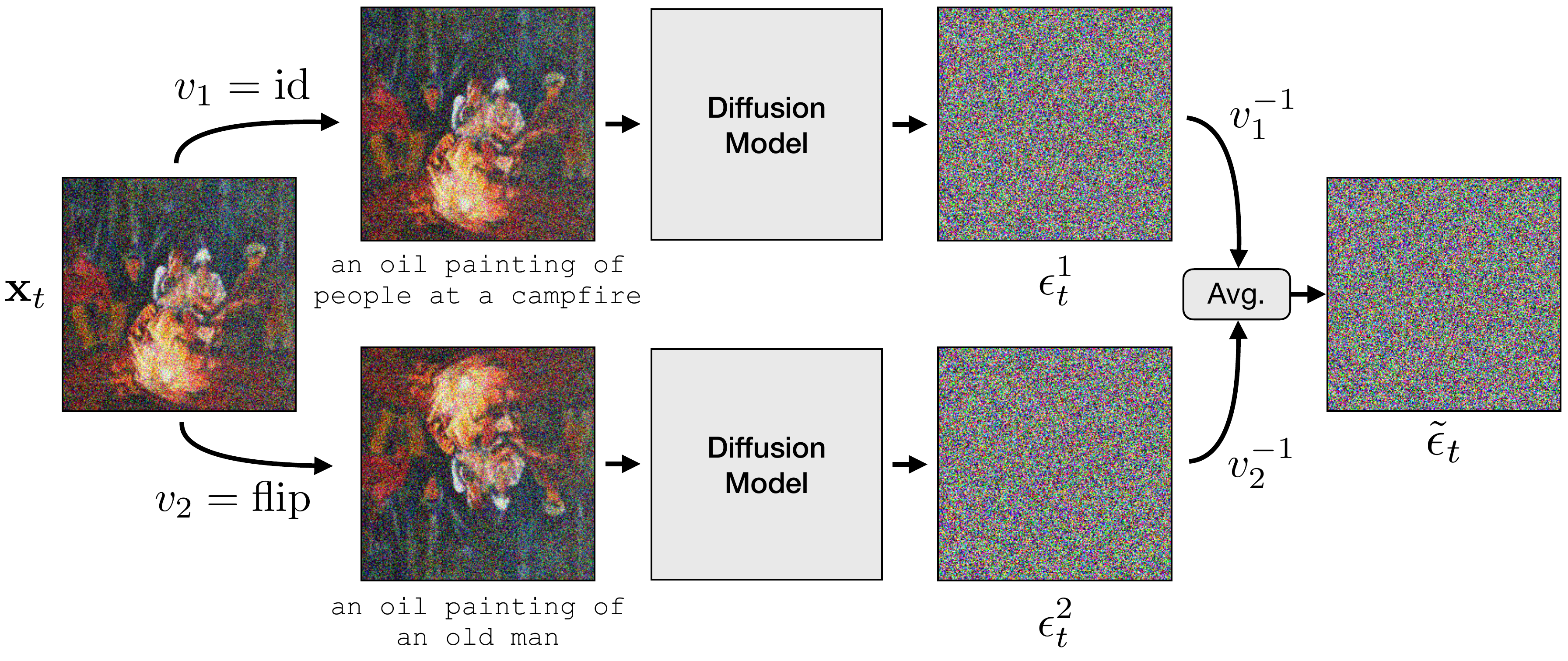}
    \caption{{\bf Algorithm Overview.} Our method works by simultaneously denoising multiple views of an image. Given a noisy image $\x_t$, we compute noise estimates, $\epsilon_t^i$, conditioned on different prompts, after applying views $v_i$. We then apply the inverse view $v_i^{-1}$ to align estimates, average the estimates, and perform a reverse diffusion step. The final output is an optical illusion.}
    \vspace{-0.5em}
\label{fig:method}
\end{figure}

Our goal is to produce multi-view optical illusions using a pretrained diffusion model. That is, we seek to synthesize images that change appearance or identity when transformed, such as when flipped or rotated.

\subsection{Text-conditioned Diffusion Models}
Diffusion models~\cite{sohldickstein2015diffusion,ho2020denoising,song2021scorebased} take i.i.d.\ Gaussian noise, $\x_T$, and iteratively denoise it to produce a sample, $\x_0$, from some data distribution. These models are parameterized by a neural network which estimates the noise in some intermediate, partially denoised data point $\x_t$, denoted as $\epsilon_\theta(\x_t, y, t)$, where $y$ is some conditioning such as text prompts and $t$ is the timestep in the diffusion process. The estimated noise is then used in an update rule~\cite{ho2020denoising,song2020denoising}, from which $\x_{t-1}$ is computed from $\x_t$.

To condition the diffusion model on another input, such as a text prompt, a common approach is to use classifier-free guidance~\cite{ho2022classifierfree}. With this method, unconditional noise estimates (usually obtained by passing the null text prompt as conditioning) and conditional noise estimates are combined together:
\begin{equation}
\label{eq:parallel}
    \epsilon_t^{\textrm{CFG}} = \epsilon_\theta(\x_t, t, \varnothing) + \gamma(\epsilon_\theta(\x_t, t, y) - \epsilon_\theta(\x_t, t, \varnothing)).
\end{equation}
Here, $\varnothing$ denotes the embedding of the empty string and $\gamma$ is a parameter that controls the strength of the guidance. Classifier-free guidance acts to sharpen the distribution of generated images to produce higher quality results. %
It also enables {\it negative prompting}~\cite{negative2022}, in which the empty text prompt embedding, 
$\varnothing$, is replaced by a text prompt that we would like to discourage the model from generating.

\setlength{\abovedisplayskip}{4pt}
\setlength{\belowdisplayskip}{4pt}

\subsection{Parallel Denoising}

We produce multi-view illusions by using a diffusion model to simultaneously denoise multiple views of an image. Concretely, we take a set of $N$ prompts, $y_i$, each associated with a view function $v_i(\cdot)$, which applies a transformation to an image. These transformations may be, for example, the identity function, an image flip, or a permutation of pixels. Then given a diffusion model, $\epsilon_\theta(\cdot)$, and a partially denoised image, $\x_t$, we combine noise estimates from different views into a single noise estimate by averaging:
\begin{equation}
    \tilde\epsilon_t = \frac{1}{N}\sum_i v_i^{-1}\left(\epsilon_\theta(v_i(\x_t), y_i, t)\right).
\end{equation}
Effectively, we use each view $v_i$ to transform the noisy image $\x_t$, estimate the noise in the transformed images, and then apply $v_i^{-1}$ to the estimates in order to transform them back to the original view. Taking an average of these noise estimates gives us our combined noise estimate, which we can then use with our choice of diffusion sampler. We note that this technique of combining noise estimates is similar to previous work on compositionality~\cite{du2020compositional,liu2021learning,liu2022compositional,du2023reduce,garipov2023compositional,du2019implicit}, where the idea is studied in further detail. In order to incorporate classifier-free guidance we simply replace the estimates $\epsilon_\theta(v_i(\x_t), y_i, t)$ with their classifier-free estimates, $\epsilon_t^{\mathrm{CFG}}$.

\subsection{Conditions on Views}
\label{sec:views}

One straightforward condition for the views is that they must be invertible. But diffusion models also implicitly impose other conditions on the views $v_i(\cdot)$. We describe two such conditions below. We find that if these conditions are not satisfied, the denoising process produces poor results.

\mypar{Linearity.} The diffusion model, $\epsilon_\theta$, acts on noisy images, $\x_t$. That is, specifically images of the form:
\begin{equation}
    \x_t = w_t^{\text{signal}}\underbrace{\x_0}_{\text{signal}} + w_t^{\text{noise}}\underbrace{\epsilon\vphantom{\x_0}}_{\text{noise}}.
\end{equation}
The exact values of $w_t^{\text{signal}}$ and $w_t^{\text{noise}}$ depend on model implementation details such as the variance schedule, but are unimportant for our work, so we omit them for clarity. What is important is that $\x_t$ is a linear combination of pure signal, $\x_0$, and pure noise, $\epsilon$, for some specific $w_t^{\text{signal}}$ and $w_t^{\text{noise}}$. Therefore our view $v_i$ must take a noisy image $\x_t$ and transform it into a new noisy image $v_i(\x_t)$ that is also a linear combination of pure signal and pure noise \textit{with the same weighting}. This can be achieved by requiring $v_i$ to be a linear transformation, of the form 
\begin{equation}
    v_i(\x_t) = \A_i\x_t,
\end{equation}
for some matrix $\A_i$, and some flattened noisy image $\x_t$. By linearity, we are effectively applying the view $v_i$ to the signal and the noise separately:
\begin{align}
    \label{eq:linear_cond}
    v_i(\x_t) &= \A_i(w_t^{\text{signal}} \x_0+w_t^{\text{noise}} \epsilon) \\
    &= w_t^{\text{signal}} \underbrace{\A_i\x_0}_{\text{new signal}} + w_t^{\text{noise}} \underbrace{\A_i\epsilon}_{\text{new noise}}.
\end{align}
This results in a linear combination of transformed signal, $\A_i\x_0$, and transformed noise, $\A_i\epsilon$, weighted with the correct scaling factors. For further discussion, please see~\cref{sec:apdx_linearity}.

\mypar{Statistical Consistency.} In addition to expecting a linear combination of signal and noise at a specific weighting, the diffusion model also expects the noise to have a precise distribution. In particular, most diffusion networks are trained with $\epsilon \sim \mathcal{N}(0, I)$. Therefore, we must ensure that our transformed noise, $\A_i\epsilon$, is likewise drawn from $\mathcal{N}(0, I)$. This is true if and only if $\A_i$ is an orthogonal matrix. We provide a proof in~\cref{sec:apdx_proof}, but intuitively this fact reflects the spherical symmetry of the standard Gaussian density. Orthogonal transformations, being generalizations of rotations and flips to higher dimensions, preserve this spherically symmetric density. Note that these are rotations {\it in the pixel values} as opposed to spatial rotations.

\subsection{Views Considered}

The vast majority of orthogonal transformations applied to an image will not correspond to an intuitive image transformation. However, a number of these transformations do. Below, we enumerate the orthogonal transformations which we consider, all of which can be seen in the illusions in~\cref{fig:teaser} unless otherwise specified.

\mypar{Identity.} The simplest transformation we consider is the identity transformation. Using this view allows us to optimize the untransformed image to align with a chosen prompt.

\mypar{Standard Image Manipulations.} We also consider \textbf{spatial rotations} of an image, which can be viewed as permutations of pixels. This works because permutations are in turn orthogonal. However, caution must be exercised when applying a rotation view, as common anti-aliasing operations such as bilinear sampling will modify the statistics of the noise. We discuss this further in~\cref{sec:failures}. \textbf{Spatial reflections} are also permutations of pixels. As such we can use these views to generate illusions. Finally, we implement an approximation to \textbf{skewing} by rolling columns of pixels by different displacements.

\mypar{General Permutations.} We have already considered the special cases of spatial rotation, reflection, and skews but we can also consider other permutations. For example, we can divide an image into jigsaw pieces and rearrange these pieces to generate jigsaw puzzles with two solutions---what we call {\bf ``polymorphic" jigsaw puzzles}. Implementation details can be found in~\cref{sec:apdx_jigsaw}.

We also consider the extreme case of sampling a {\bf completely random permutation} of pixels and treating it as our view. Additionally, we can reduce the complexity of this by considering {\bf permutations of square patches}, rather than pixels. Examples of these illusions can be found in~\cref{fig:permutations} and are discussed in~\cref{sec:qualitative}.

Finally, we consider rotating a circle within an image while leaving the rest of the image stationary, which we term {\bf inner rotations}. Note that the permutations we consider are certainly not exhaustive, and many clever transformations exist which we do not study.

\mypar{Color Inversion.} Negation is an orthogonal transform; it is intuitively a 180 degree rotation generalized to higher dimensions. This allows us to generate illusions that change appearance upon color inversion, assuming pixel values are centered at 0 (\eg, in the range [-1, 1]).

\mypar{Arbitrary Orthogonal Transformations.} An arbitrary rotation of an image in pixel space is uninterpretable. Nevertheless, we demonstrate that our method works for these transformations as well. While these ``illusions" are inscrutable to the human eye, they serve as confirmation that any orthogonal transformation works as a view with our method. These can be found in~\cref{fig:orthogonal} and are discussed in~\cref{sec:qualitative}.

\subsection{Design Decisions}

Beyond the core method, we also consider various design decisions aimed at maximizing illusion quality.

\begin{figure}[t]
    \centering
    \includegraphics[width=\linewidth]{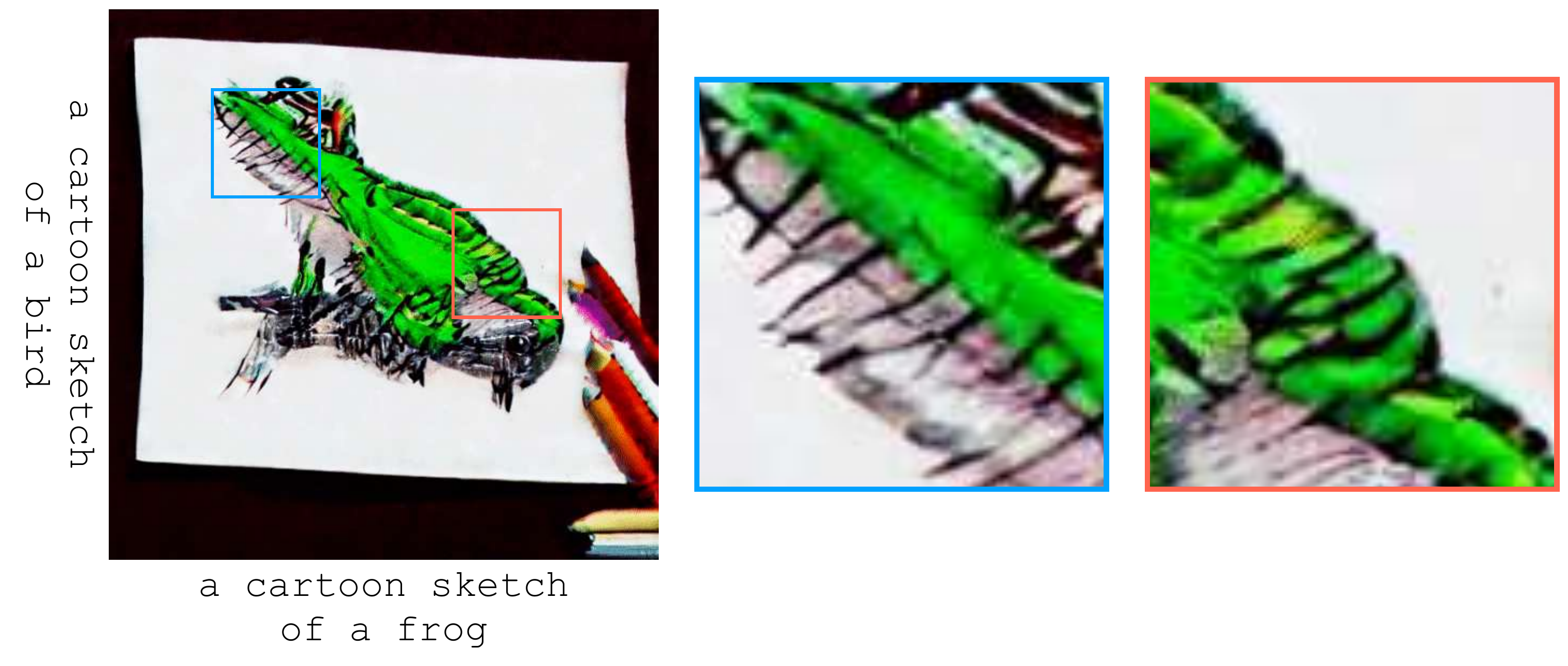}
    \caption{{\bf Latent-Based Artifacts.} Manipulating the {\it location} of latent codes does not change the {\it orientation} of the blocks for which they encode. Therefore, when using latent diffusion models we see artifacts as shown above, in which straight lines are thatched under a rotation.}
    \vspace{-1em}
\label{fig:latent}
\end{figure}

\mypar{Pixel Diffusion Model.} Previous work~\cite{tancik2023illusions} performed multi-view denoising using Stable Diffusion~\cite{rombach2022ldm}, a latent diffusion model. However, the latent representation effectively encodes patches of pixels. This leads to artifacts under rotations or flips, where the {\it location} of latents change, but the {\it content} and {\it orientation} of these blocks do not. We show a qualitative example of this in~\cref{fig:latent}, in which the model is forced to generate thatched lines to produce straight lines under a $90^\circ$ rotation.

To ameliorate this issue, we implement our method using a pixel-based diffusion model, DeepFloyd IF~\cite{deepfloyd2023}. DeepFloyd denoises directly on pixels, effectively side-stepping the problem of orientation in latent code blocks.

\mypar{Combining Noise Estimates.} In addition to taking a mean of noise estimates from different views, we also consider alternating through them by timestep, using the estimate
\begin{equation}
    \tilde\epsilon_t = v_{t\ \mathrm{mod}\ N}^{-1} \left( \epsilon_\theta(v_{t\ \mathrm{mod}\ N}(\x_t), t, y) \right).
\end{equation}
This is the reduction strategy used by~\cite{tancik2023illusions}, but we show in ablations in~\cref{sec:ablations} that it performs worse than averaging.

\mypar{Negative Prompting.} We experiment with negative prompting~\cite{negative2022} in the 2-view case by using one view's prompt as a negative for the other view, and vice versa. This encourages the model to hide the other view's prompt for a given view. For a discussion, please see the ablations in~\cref{sec:ablations}.

\section{Results}

We provide quantitative and qualitative results, and quantitative ablations. If not specified, qualitative results have been hand picked for quality. For {\bf random samples} please see \cref{fig:random} and~\cref{sec:apdx_random}. All implementation details can be found in~\cref{sec:apdx_impl}.

\begin{table}[t!] %
    \begin{center}
    \caption{{\bf Quantitative Results.} We report the alignment score, $\mathcal{A}$, and the concealment score, $\mathcal{C}$, as well as quantiles of these scores. For a discussion, please see \cref{sec:quant}.}
    
    \label{tbl:quantitative}
    \setlength\tabcolsep{3pt}
    \resizebox{\linewidth}{!}{
    \begin{tabular}{llcccccc}
    \toprule
    Prompt Pair & Method
    & $\mathcal{A}\uparrow$ & $\mathcal{A}_{0.9}\uparrow$ & $\mathcal{A}_{0.95}\uparrow$ & $\mathcal{C}\uparrow$ & $\mathcal{C}_{0.9}\uparrow$ & $\mathcal{C}_{0.95}\uparrow$ \\
    \midrule
        \multirow{ 3}{*}{CIFAR} & Burgert~\etal~\cite{burgert2023illusions} & 0.225 & 0.253 & 0.260 & 0.501 & 0.526 & 0.537 \\
    & Tancik~\cite{tancik2023illusions} & 0.278 & 0.310 & 0.316 & 0.595 & 0.692 & 0.712 \\
    & Ours & \textbf{0.287} & \textbf{0.321} & \textbf{0.327} & \textbf{0.624} & \textbf{0.717} & \textbf{0.739} \\
    \cmidrule(r){1-8}
    \multirow{ 3}{*}{Ours} & Burgert~\etal~\cite{burgert2023illusions} & 0.233 & 0.270 & 0.283 & 0.501 & 0.526 & 0.538 \\
    & Tancik~\cite{tancik2023illusions} & 0.256 & 0.294 & 0.309 & 0.545 & 0.621 & 0.655 \\
    & Ours & \textbf{0.275} & \textbf{0.315} & \textbf{0.326} & \textbf{0.574} & \textbf{0.668} & \textbf{0.694} \\
    
    \bottomrule
    \end{tabular}
   }
    \end{center}
    \vspace{-4mm}
\end{table}
\definecolor{ours}{HTML}{d1ebd7}
\begin{table}[t!]
    \begin{center}
    \caption{{\bf Ablations.} We ablate negative prompting, reduction methods, and guidance scales on our dataset.}
    
    \label{tbl:ablations}
    \setlength\tabcolsep{3pt}
    \resizebox{\linewidth}{!}{
    \begin{tabular}{lcccccc}
    \toprule
     
    Ablation & $\mathcal{A}\uparrow$ & $\mathcal{A}_{0.9}\uparrow$ & $\mathcal{A}_{0.95}\uparrow$ & $\mathcal{C}\uparrow$ & $\mathcal{C}_{0.9}\uparrow$ & $\mathcal{C}_{0.95}\uparrow$ \\
    \midrule
    Negative Prompting & 0.24 & 0.27 & 0.276 & \textbf{0.576} & \textbf{0.659} & \textbf{0.683} \\
    No Negative Prompting & \textbf{0.255} & \textbf{0.285} & \textbf{0.295} & 0.567 & 0.643 & 0.679 \\
    \cmidrule(r){1-7} 
    
    Alternating Reduction & 0.252 & \textbf{0.286} & 0.292 & 0.560 & 0.639 & 0.664 \\
    Mean Reduction & \textbf{0.255} & 0.285 & \textbf{0.295} & \textbf{0.567} & \textbf{0.643} & \textbf{0.679} \\
    \cmidrule(r){1-7} 
    
    $\gamma = 3.0$ & 0.239 & 0.271 & 0.285 & 0.537 & 0.610 & 0.629 \\
    $\gamma = 7.0$ & 0.255 & 0.285 & 0.295 & 0.567 & 0.643 & 0.679 \\
    $\gamma = 10.0$ & \textbf{0.259} & \textbf{0.290} & \textbf{0.297} & \textbf{0.576} & \textbf{0.664} & \textbf{0.702} \\
    
    \bottomrule
    \end{tabular}
   }
    \end{center}
    \vspace{-4mm}
\end{table}

\subsection{Quantitative Results}
\label{sec:quant}
\vspace{1em}
\mypar{Metrics.} We use CLIP~\cite{radford2021clip} to measure how well views align with the desired prompts. We consider two metrics derived from a score matrix $\mathbf{S} \in \mathbb{R}^{N\times N}$, defined as
\begin{equation}
    \mathbf{S}_{ij} = \phi_{\text{img}}(v_i(\x))^T\phi_{\text{text}}(p_j),
\end{equation}
where $\phi_{\text{img}}$ and $\phi_{\text{text}}$ are the CLIP visual and textual encoders respectively, returning a unit-norm vector embedding. $\x$ is our generated illusion, and $v_i$ are our views with associated prompts $p_i$. A higher dot product indicates higher similarity between the image and text.

The first metric we consider is $\min\operatorname{diag}(\mathbf{S})$, which intuitively measures the worst alignment of all the views. We term this metric $\mathcal{A}$, the {\bf alignment score}. However, this metric does not account for the possibility of seeing prompt $p_i$ in view $v_j$ for $i\neq j$. This is an occasional failure case of our method and to quantify this we propose a second derived metric which we term $\mathcal{C}$, the {\bf concealment score}, computed as
\begin{equation}
\frac{1}{N}\operatorname{tr}(\operatorname{softmax}(S/\tau)),
\end{equation}
where $\tau$ is the temperature parameter of CLIP. In computing this metric we average both directions of the softmax, so that this metric measures how well CLIP can classify a view as one of the $N$ prompts and vice versa.

\begin{figure}[t]
    \centering
    \includegraphics[width=\linewidth]{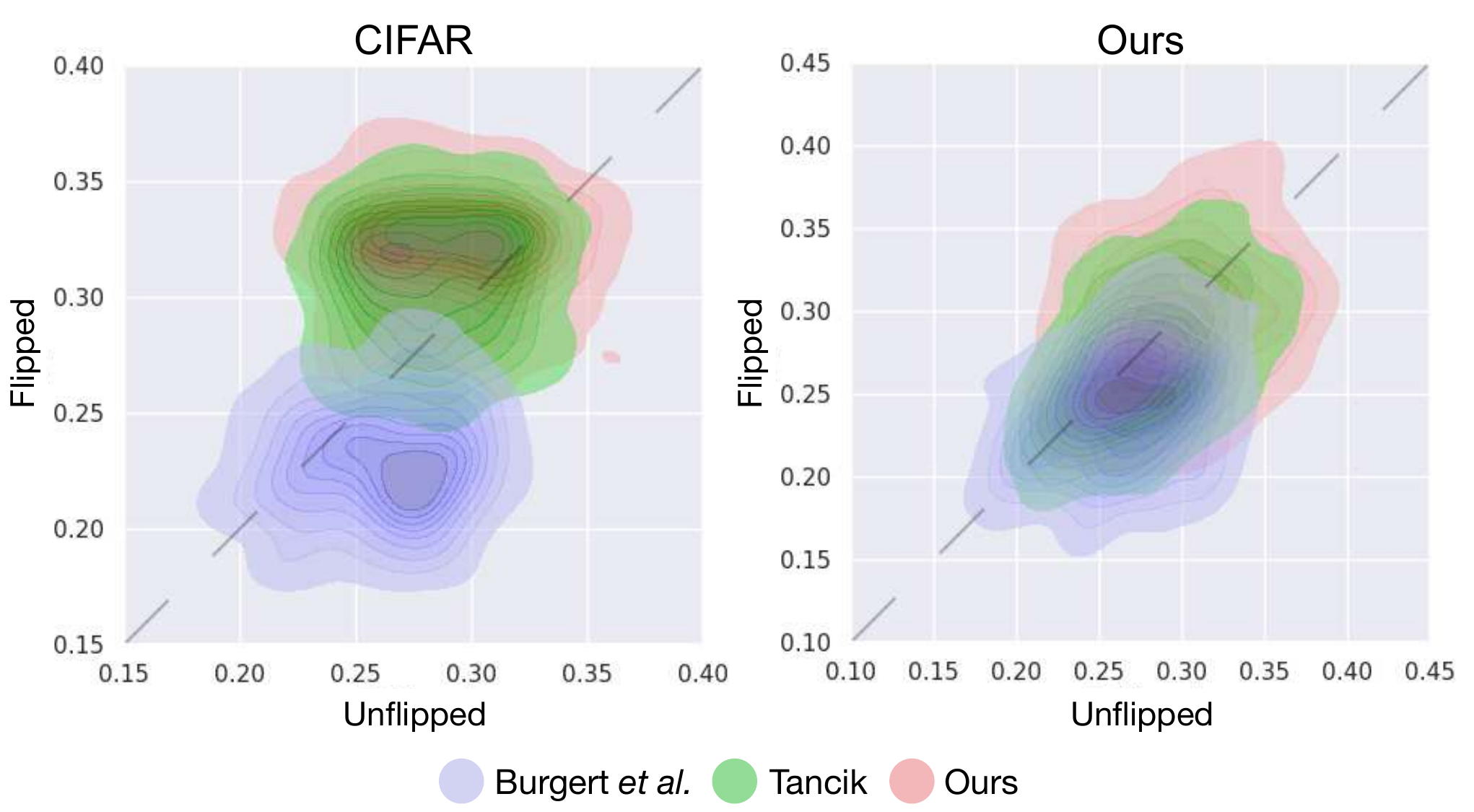}
    \caption{{\bf Flip View CLIP Score Distribution.} We visualize trade-offs between flipped and unflipped views by plotting the distribution of CLIP scores on the datasets. Note that the quality of the flipped image is as good as the unflipped image, with parity indicated by the dashed line.}
\label{fig:density}
\end{figure}

\mypar{Dataset.} To evaluate our method and baselines we compile two datasets of prompt pairs for 2-view illusions. One dataset uses the 10 classes from CIFAR-10 and contains a prompt per pair of classes, for a total of 45 prompt pairs. We refer to this as {\bf CIFAR}. The other dataset we compile by hand, with the process documented in~\cref{sec:apdx_dataset}. This dataset consists of 50 prompt pairs, which we refer to as {\bf Ours}.

\mypar{Baselines.} We use two baselines that generate illusions using off-the-shelf diffusion models. One, which we denote ``Burgert~\etal~\cite{burgert2023illusions}," uses Score Distillation Sampling. The other, which we denote ``Tancik~\cite{tancik2023illusions}," is an earlier version of our method, with differences discussed in more detail in~\cref{sec:related_work}

\mypar{Results.} We show results comparing our method to baselines on both datasets in \cref{tbl:quantitative} using vertical flips. We use vertical flips because it is a transformation supported by our method as well as the baselines. We use 10 samples per prompt, for a total of 450 and 500 samples for the CIFAR dataset and our dataset respectively. It is hard to perform a fair comparison with more samples because the Burgert~\etal method uses SDS, which is quite slow\footnote{Sampling just 10 images per prompt already takes more than a week of GPU-hours.}. Because we are particularly interested in the ``best-case" performance, we also report quantiles of metrics, which we denote as $\mathcal{A}_{0.9}$ for the 90th percentile, for example. As can be seen, our method performs consistently better than the baselines, in both the alignment score and the concealment score. 

In order to give a clearer understanding of trade-offs when optimizing two views, we show density plots which plot the CLIP scores of each of the two views of an illusion in~\cref{fig:density}. As can be seen, we do better than the baselines on average and in the best-case. Moreover, flipping during denoising does not hurt performance. The quality of the flipped images is as high as the unflipped images.

\subsection{Ablations}
\label{sec:ablations}

We ablate out the noise estimate reduction strategy, negative prompting, and the guidance scale in~\cref{tbl:ablations}. We use our dataset, with 10 samples for each prompt for a total of 500 illusions.

\mypar{Reduction Strategy.} We find that mean reduction does better than alternating. Our hypothesis is that alternating the noise estimates results in ``thrashing," causing poor convergence. Moreover, we find that the alternating strategy gives poor results on illusions with more than 2 views, as each view has fewer denoising steps. Qualitative examples of this can be found in~\cref{sec:apdx_reduction}.

\begin{figure}[t]
    \centering
    \includegraphics[width=\linewidth]{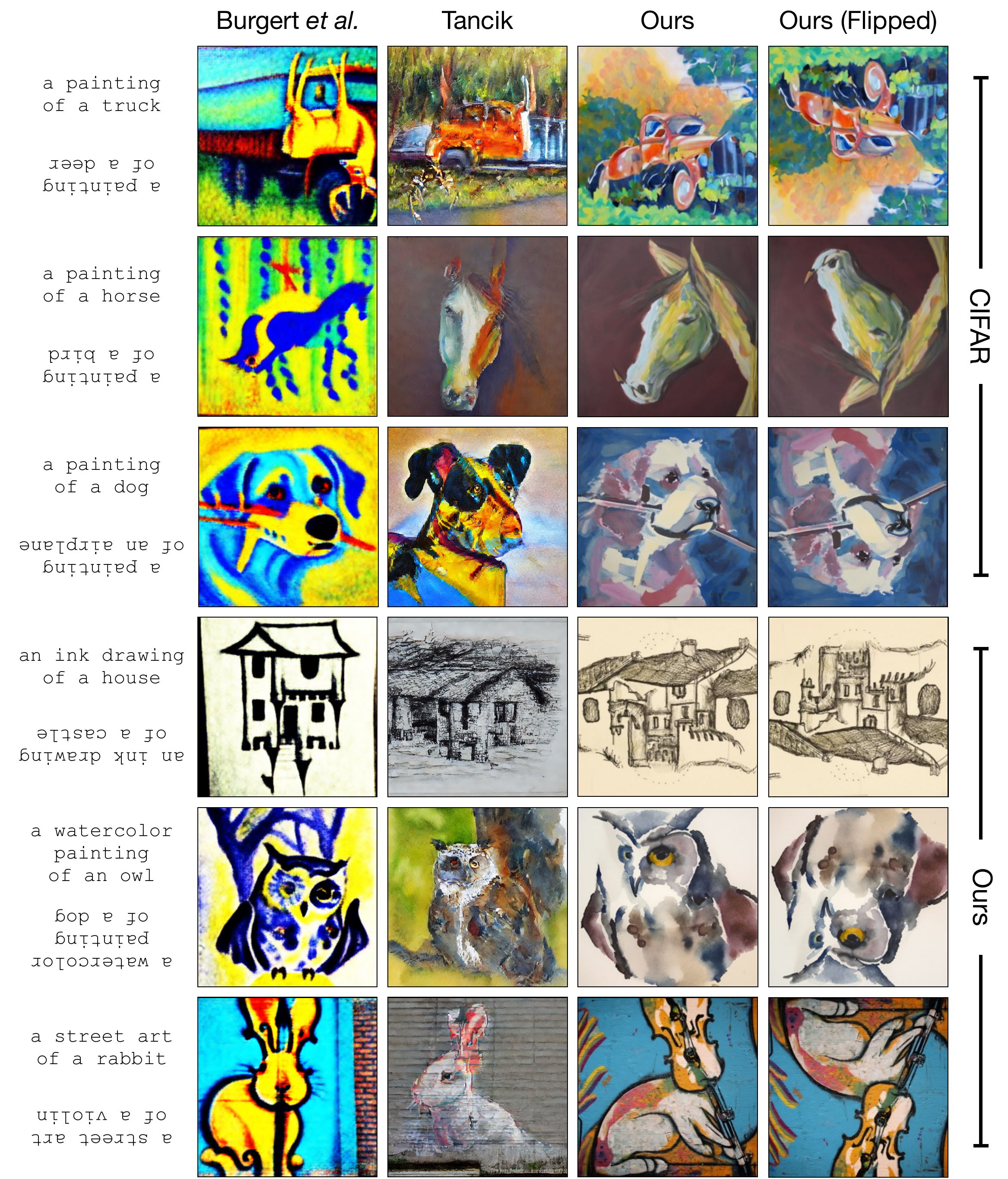}
    \caption{{\bf Qualitative Comparisons.} We compare illusions generated by baselines to our illusions. We show examples from both our prompt dataset and the CIFAR prompt dataset.}
    \vspace{-1em}
\label{fig:qualitative}
\end{figure}

\mypar{Negative Prompting.} When using negative prompting, care must be taken to omit any overlap between the negative and positive prompt. For example, given the two prompts \texttt{"oil painting of a dog"} and \texttt{"oil painting of a cat"}, using one prompt as the negative for the other would simultaneously encourage and discourage the style \texttt{"oil painting"}. Rather, the negative prompts should be \texttt{"a cat"} and \texttt{"a dog"} respectively. We find that negative prompting can improve the concealment score, indicating that it is working as intended. But this comes at the cost of worse alignment score. This is because the negative and positive prompt may have fundamental similarities. For example, using \texttt{"a cat"} as the negative prompt for the prompt \texttt{"an oil painting of a dog"} may discourage the model from synthesizing anything remotely cat-like---such as fur, four legs, or a tail---even if it helps in producing a dog. For this reason we opt not to use negative prompting with our method.

\mypar{Guidance Scale.} We also ablate out various guidance scales, $\gamma$, for our method. We find that a higher guidance scale tends to do better. This is presumably because a higher guidance scale results in a sharper sampling distribution.

\subsection{Qualitative Results}
\label{sec:qualitative}

\begin{figure}[t]
    \centering
    \includegraphics[width=\linewidth]{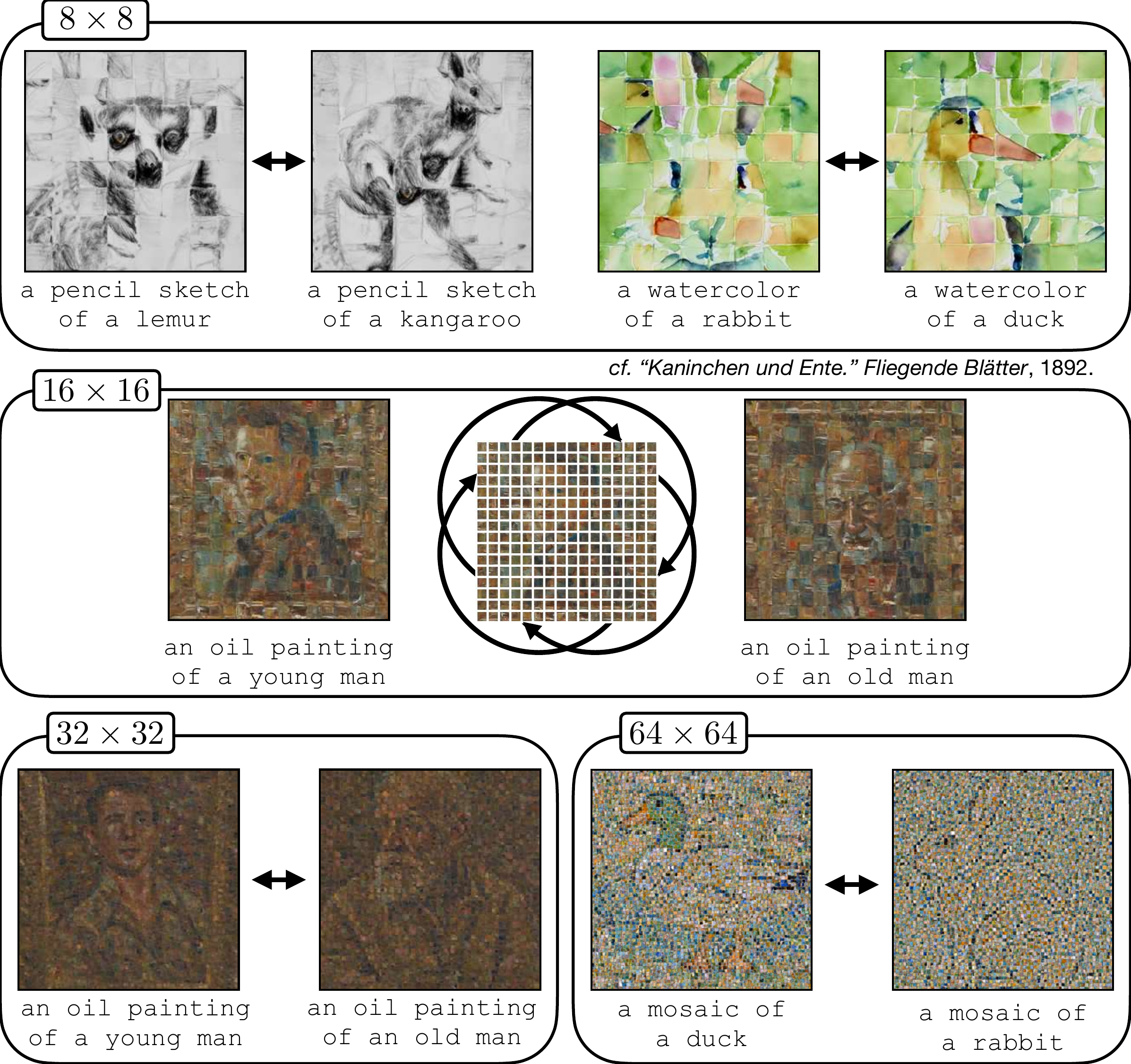}
    \caption{{\bf Permutation Illusions.} We synthesize images whose appearance changes upon permutation of patches. Even in the difficult case of a $64\times 64$ grid of patches, in which every pixel is effectively shuffled, we are able to generate meaningful images.}
    \vspace{-1em}
\label{fig:permutations}
\end{figure}

\begin{figure}[t]
    \centering
    \includegraphics[width=\linewidth]{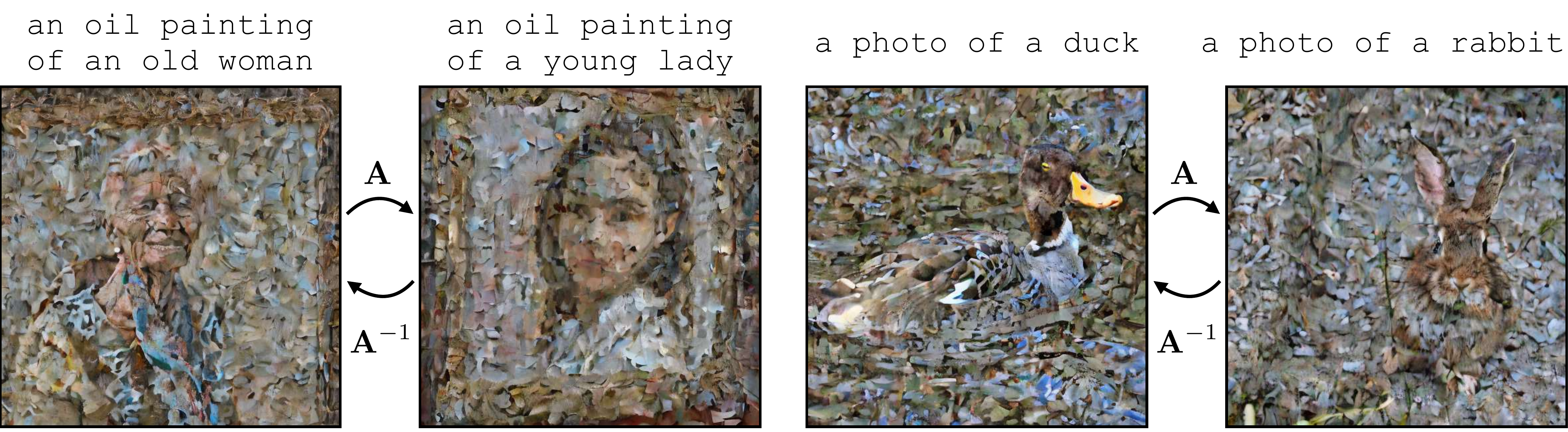}
    \caption{{\bf Orthogonal Illusions.} We show that our method works, even when the view is a randomly sampled orthogonal transformation $\A$. While these ``illusions" are incomprehensible to human perception, they serve as a confirmation for our mathematical analysis.}
    \vspace{-5mm}
\label{fig:orthogonal}
\end{figure}

\begin{figure*}[t]
    \centering
    \includegraphics[width=\linewidth]{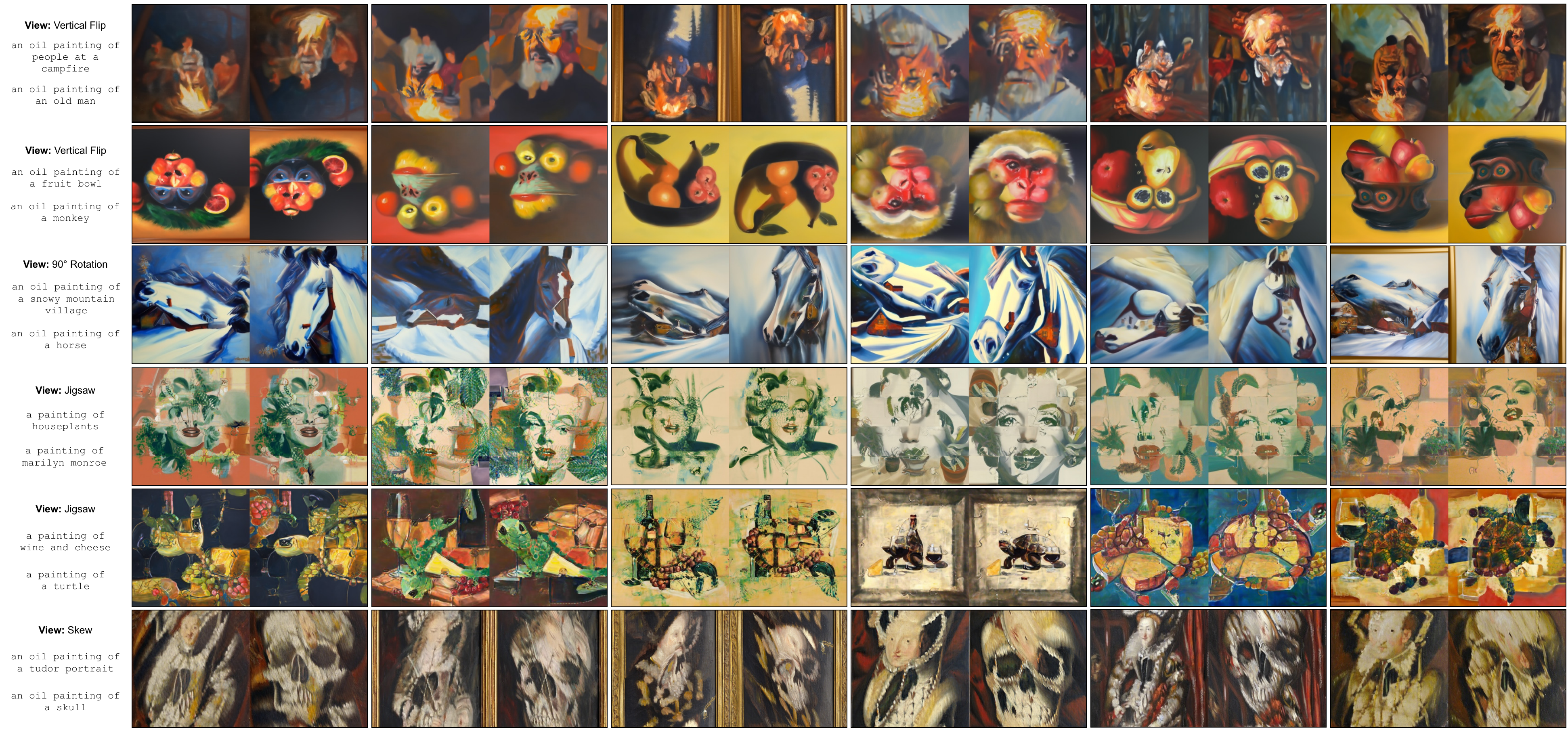}
    \caption{{\bf Random Samples.} We show random samples, along with their corresponding view, for selected prompts. For more random samples please see~\cref{sec:apdx_random}. \textbf{For best quality, view digitally and zoom-in.}}
    \vspace{-1em}
\label{fig:random}
\end{figure*}

We show qualitative results in \cref{fig:teaser}, \cref{fig:qualitative}, \cref{fig:permutations}, and \cref{fig:orthogonal}. Again, random samples may be found in \cref{fig:random} and~\cref{sec:apdx_random}. Additional qualitative samples can be found in~\cref{sec:apdx_results}. Overall, we find that our method can produce very high quality optical illusions for a wide range of views. Interestingly, our method often finds clever ways of reusing elements from one view for another, such as in the \texttt{"waterfalls"/"rabbit"/"teddy bear"} three-view illusion in \cref{fig:teaser}, in which the nose of the teddy bear is the eye of the rabbit, and a rock on the waterfalls.

\mypar{Baselines.} We provide qualitative comparisons of our method to baselines in~\cref{fig:qualitative}, where we pick the best images out of 100 samples for each method. As can be seen, images generated using our method match the prompts in both views equally well and are of higher quality.

\mypar{Permutations.} Pixel and patch permutations, being a subset of orthogonal transformations, should work with our method. We show that this is indeed the case in \cref{fig:permutations}, where we have results on patch grids of various sizes under randomly sampled permutations. The $64\times 64$ case is quite hard, yet our method is able to generate images that satisfy the constraint, albeit at lower quality.

\mypar{Arbitrary Orthogonal Transformations.} As discussed in \cref{sec:views}, our method works for any orthogonal transformation. So far, we have shown illusions based on a subset of orthogonal views that correspond to intuitive image transformations. In \cref{fig:orthogonal}, we show ``illusions" using an arbitrary orthogonal transformation as a view. We use Stable Diffusion~\cite{rombach2022ldm} and sample a random orthogonal matrix $\A\in\mathbb{R}^{16384\times16384}$ by projecting an i.i.d. random Gaussian matrix with an SVD. These dimensions correspond to the size of the Stable Diffusion latent space. We note that this is an incredibly hard and unnatural transformation of an image, and results are accordingly of lower quality, but our method is still able to produce reasonable images. 

\mypar{Random Samples.} We show random samples for selected prompts in \cref{fig:random}. As can be seen, these random samples, while not as good as those in \cref{fig:teaser}, are still very high quality. Some failure cases can be seen where the model prefers one prompt over another. We add further discussion and present more random samples in~\cref{sec:apdx_random}.

\subsection{Failures}
\label{sec:failures}

\begin{figure}[t]
    \centering
    \includegraphics[width=\linewidth]{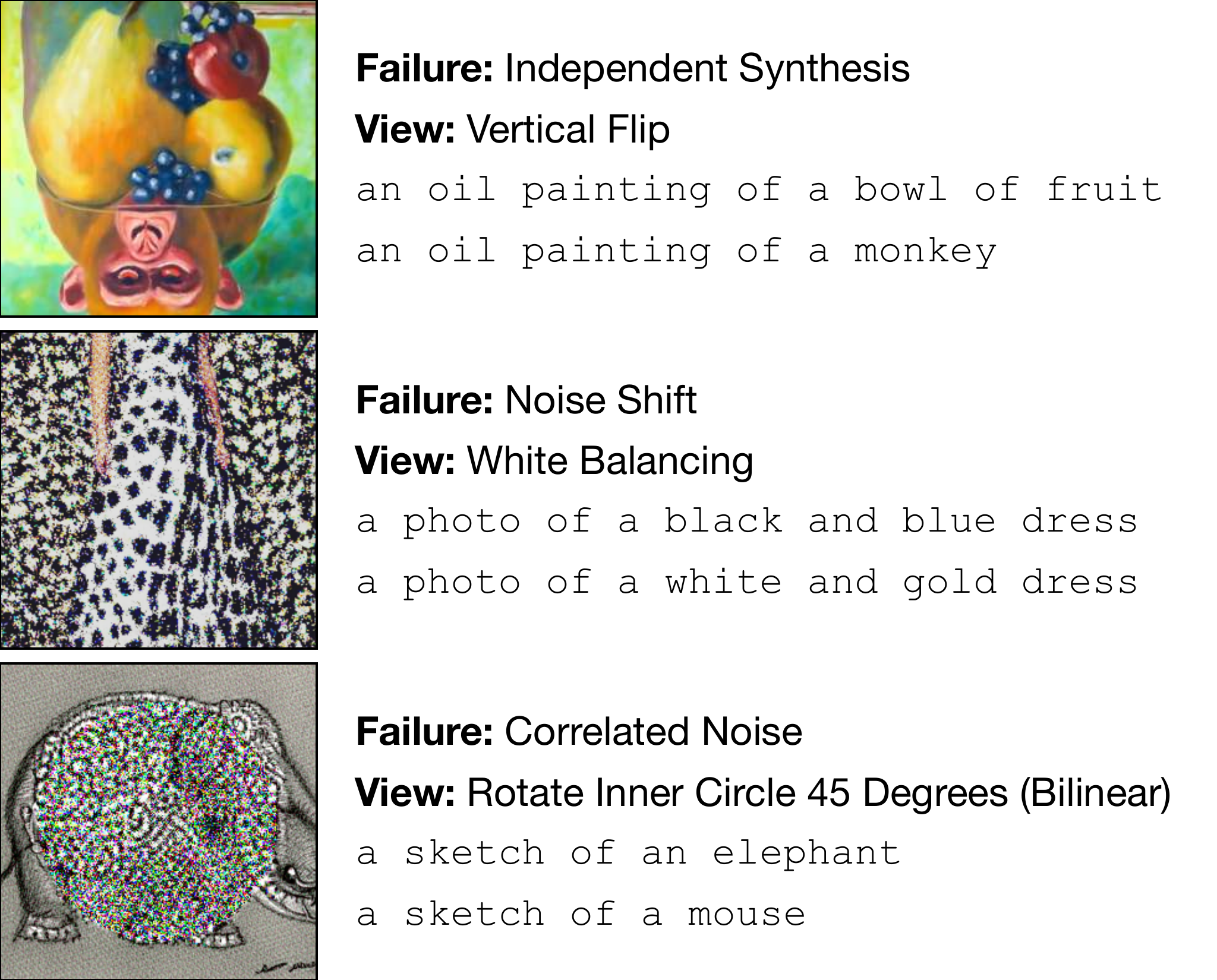}
    \caption{{\bf Failures.} We highlight three interesting failure cases, which are discussed in \cref{sec:failures}.}
    \vspace{-5mm}
\label{fig:failure}
\end{figure}

We highlight three interesting failure cases of our method in~\cref{fig:failure}. 

\mypar{Independent Synthesis.} The first of these cases involves the model synthesizing prompts separately, without combining elements of the two to form an illusion. Empirically, this happens surprisingly rarely, especially given that it seems to be such an easy shortcut solution. We hypothesize that this is because the diffusion model is biased toward centering its content, resulting in far more images with content that is integrated and centered as opposed to separate and off-center.

\mypar{Noise Shift.} Using views that preserve noise statistics is critical to our method's success. For example, we attempted to recreate the ``Dress" illusion~\cite{dress_wikipedia}, in which a dress can be seen as either ``blue and black" or ``white and gold." We used simple white balancing as our view, in which pixel values were scaled by a constant factor. While this transformation is linear, it does not preserve the statistics of Gaussian noise. As a result, we see artifacts in the forms of spots, which we hypothesize is the result of the model interpreting the scaled Gaussian noise as signal and actively denoising peaks in the scaled noise.

\mypar{Correlated Noise.} While our method supports rotations as transformations, as demonstrated with the ``3-view," ``4-view," and ``Inner Rotation" illusions in~\cref{fig:teaser}, care must be taken that the rotation does not introduce correlations in the noise, such as through anti-aliasing. For example, bilinear sampling introduces significant correlations in the noise, as it is a linear combination of four adjacent pixels. Therefore, seemingly innocuous rotations may result in divergent samples if transformations are not carefully kept correlation free, as shown with the 45 degree bilinear rotation in~\cref{fig:failure}.

\section{Limitations and Conclusions}

We present a method to produce compelling and diverse optical illusions. Our method is simple and straightforward to implement, and additionally amenable to theoretical analysis. We prove that our method works for a broad set of transformations, and qualitatively show that it can generate a wide array of optical illusions. However, at the same time many possible illusions and transformations are still not possible using our method, such as color constancy illusions, homographies, stretches, and more generally non-volume-preserving deformations. We leave implementation of these views for future work. Moreover, our method does not consistently produce perfect illusions. This may be a symptom of the difficulty of producing good illusions, but may indicate future work to be done to improve consistency.

\mypar{Acknowledgements} We thank William Henning, Trenton Chang, Kimball Strong, Jeongsoo Park, Patrick Chao, Kurtland Chua, and Mohamed El Banani for their feedback on early drafts. Daniel is supported by the National Science Foundation Graduate Research Fellowship under Grant No. 1841052.
{
    \vspace{-2em}   
    \small
    \bibliographystyle{ieeenat_fullname}
    \bibliography{main}

\begin{thebibliography}{47}
\providecommand{\natexlab}[1]{#1}
\providecommand{\url}[1]{\texttt{#1}}
\expandafter\ifx\csname urlstyle\endcsname\relax
  \providecommand{\doi}[1]{doi: #1}\else
  \providecommand{\doi}{doi: \begingroup \urlstyle{rm}\Url}\fi

\bibitem[AUTOMATIC1111(2022)]{negative2022}
AUTOMATIC1111.
\newblock Negative prompt.
\newblock
  \url{https://github.com/AUTOMATIC1111/stable-diffusion-webui/wiki/Negative-prompt},
  2022.
\newblock Accessed: November 7, 2023.

\bibitem[Burgert et~al.(2023)Burgert, Li, Leite, Ranasinghe, and
  Ryoo]{burgert2023illusions}
Ryan Burgert, Xiang Li, Abe Leite, Kanchana Ranasinghe, and Michael Ryoo.
\newblock Diffusion illusions: Hiding images in plain sight.
\newblock \url{https://ryanndagreat.github.io/Diffusion-Illusions}, 2023.

\bibitem[Chandra et~al.(2022)Chandra, Li, Tenenbaum, and
  Ragan-Kelley]{chandra2022designing}
Kartik Chandra, Tzu-Mao Li, Joshua Tenenbaum, and Jonathan Ragan-Kelley.
\newblock Designing perceptual puzzles by differentiating probabilistic
  programs.
\newblock In \emph{ACM SIGGRAPH 2022 Conference Proceedings}, pages 1--9, 2022.

\bibitem[Chi et~al.(2014)Chi, Yao, Zhang, and Lee]{chi2014optical}
Ming-Te Chi, Chih-Yuan Yao, Eugene Zhang, and Tong-Yee Lee.
\newblock Optical illusion shape texturing using repeated asymmetric patterns.
\newblock \emph{The Visual Computer}, 30:\penalty0 809--819, 2014.

\bibitem[Chu et~al.(2010)Chu, Hsu, Mitra, Cohen-Or, Wong, and
  Lee]{chu2010camouflage}
Hung-Kuo Chu, Wei-Hsin Hsu, Niloy~J Mitra, Daniel Cohen-Or, Tien-Tsin Wong, and
  Tong-Yee Lee.
\newblock Camouflage images.
\newblock \emph{ACM Trans. Graph.}, 29\penalty0 (4):\penalty0 51--1, 2010.

\bibitem[Dhariwal and Nichol(2021)]{dhariwal2021diffusion}
Prafulla Dhariwal and Alexander Nichol.
\newblock Diffusion models beat gans on image synthesis.
\newblock \emph{Advances in neural information processing systems},
  34:\penalty0 8780--8794, 2021.

\bibitem[Du and Mordatch(2019)]{du2019implicit}
Yilun Du and Igor Mordatch.
\newblock Implicit generation and generalization in energy-based models.
\newblock \emph{arXiv preprint arXiv:1903.08689}, 2019.

\bibitem[Du et~al.(2020)Du, Li, and Mordatch]{du2020compositional}
Yilun Du, Shuang Li, and Igor Mordatch.
\newblock Compositional visual generation with energy based models.
\newblock \emph{Advances in Neural Information Processing Systems},
  33:\penalty0 6637--6647, 2020.

\bibitem[Du et~al.(2023)Du, Durkan, Strudel, Tenenbaum, Dieleman, Fergus,
  Sohl-Dickstein, Doucet, and Grathwohl]{du2023reduce}
Yilun Du, Conor Durkan, Robin Strudel, Joshua~B Tenenbaum, Sander Dieleman, Rob
  Fergus, Jascha Sohl-Dickstein, Arnaud Doucet, and Will~Sussman Grathwohl.
\newblock Reduce, reuse, recycle: Compositional generation with energy-based
  diffusion models and mcmc.
\newblock In \emph{International Conference on Machine Learning}, pages
  8489--8510. PMLR, 2023.

\bibitem[Ehm(2011)]{ehm2011variational}
Werner Ehm.
\newblock A variational approach to geometric-optical illusions modeling.
\newblock \emph{Proceedings of Fechner Day}, 27\penalty0 (1):\penalty0 41--46,
  2011.

\bibitem[Elsayed et~al.(2018)Elsayed, Shankar, Cheung, Papernot, Kurakin,
  Goodfellow, and Sohl-Dickstein]{elsayed2018adversarial}
Gamaleldin Elsayed, Shreya Shankar, Brian Cheung, Nicolas Papernot, Alexey
  Kurakin, Ian Goodfellow, and Jascha Sohl-Dickstein.
\newblock Adversarial examples that fool both computer vision and time-limited
  humans.
\newblock \emph{Advances in neural information processing systems}, 31, 2018.

\bibitem[Freeman et~al.(1991)Freeman, Adelson, and Heeger]{freeman1991motion}
William~T Freeman, Edward~H Adelson, and David~J Heeger.
\newblock Motion without movement.
\newblock \emph{ACM Siggraph Computer Graphics}, 25\penalty0 (4):\penalty0
  27--30, 1991.

\bibitem[Garipov et~al.(2023)Garipov, De~Peuter, Yang, Garg, Kaski, and
  Jaakkola]{garipov2023compositional}
Timur Garipov, Sebastiaan De~Peuter, Ge Yang, Vikas Garg, Samuel Kaski, and
  Tommi Jaakkola.
\newblock Compositional sculpting of iterative generative processes.
\newblock \emph{arXiv preprint arXiv:2309.16115}, 2023.

\bibitem[Gomez-Villa et~al.(2019)Gomez-Villa, Martin, Vazquez-Corral, and
  Bertalm{\'\i}o]{gomez2019convolutional}
Alexander Gomez-Villa, Adrian Martin, Javier Vazquez-Corral, and Marcelo
  Bertalm{\'\i}o.
\newblock Convolutional neural networks can be deceived by visual illusions.
\newblock In \emph{Proceedings of the IEEE/CVF conference on computer vision
  and pattern recognition}, pages 12309--12317, 2019.

\bibitem[Gomez-Villa et~al.(2022)Gomez-Villa, Mart{\'\i}n, Vazquez-Corral,
  Bertalm{\'\i}o, and Malo]{gomez2022synthesis}
Alex Gomez-Villa, Adri{\'a}n Mart{\'\i}n, Javier Vazquez-Corral, Marcelo
  Bertalm{\'\i}o, and Jes{\'u}s Malo.
\newblock On the synthesis of visual illusions using deep generative models.
\newblock \emph{Journal of Vision}, 22\penalty0 (8):\penalty0 2--2, 2022.

\bibitem[Goodfellow et~al.(2014)Goodfellow, Shlens, and
  Szegedy]{goodfellow2014explaining}
Ian~J Goodfellow, Jonathon Shlens, and Christian Szegedy.
\newblock Explaining and harnessing adversarial examples.
\newblock \emph{arXiv preprint arXiv:1412.6572}, 2014.

\bibitem[Graikos et~al.(2022)Graikos, Malkin, Jojic, and
  Samaras]{graikos2022diffusion}
Alexandros Graikos, Nikolay Malkin, Nebojsa Jojic, and Dimitris Samaras.
\newblock Diffusion models as plug-and-play priors.
\newblock \emph{Advances in Neural Information Processing Systems},
  35:\penalty0 14715--14728, 2022.

\bibitem[Guo et~al.(2023)Guo, Collins, de~Lima, and Owens]{guo2023ganmouflage}
Rui Guo, Jasmine Collins, Oscar de Lima, and Andrew Owens.
\newblock Ganmouflage: 3d object nondetection with texture fields.
\newblock \emph{Computer Vision and Pattern Recognition (CVPR)}, 2023.

\bibitem[Hertzmann(2020)]{hertzmann2020visual}
Aaron Hertzmann.
\newblock Visual indeterminacy in gan art.
\newblock In \emph{ACM SIGGRAPH 2020 Art Gallery}, pages 424--428. 2020.

\bibitem[Hirsch and Tal(2020)]{hirsch2020color}
Elad Hirsch and Ayellet Tal.
\newblock Color visual illusions: A statistics-based computational model.
\newblock \emph{Advances in neural information processing systems},
  33:\penalty0 9447--9458, 2020.

\bibitem[Ho and Salimans(2022)]{ho2022classifierfree}
Jonathan Ho and Tim Salimans.
\newblock Classifier-free diffusion guidance, 2022.

\bibitem[Ho et~al.(2020)Ho, Jain, and Abbeel]{ho2020denoising}
Jonathan Ho, Ajay Jain, and Pieter Abbeel.
\newblock Denoising diffusion probabilistic models.
\newblock \emph{arXiv preprint arxiv:2006.11239}, 2020.

\bibitem[Jaini et~al.(2023)Jaini, Clark, and Geirhos]{jaini2023intriguing}
Priyank Jaini, Kevin Clark, and Robert Geirhos.
\newblock Intriguing properties of generative classifiers.
\newblock \emph{arXiv preprint arXiv:2309.16779}, 2023.

\bibitem[Konstantinov et~al.(2023)Konstantinov, Shonenkov, Bakshandaeva, and
  Ivanova]{deepfloyd2023}
Mikhail Konstantinov, Alex Shonenkov, Daria Bakshandaeva, and Ksenia Ivanova.
\newblock If by deepfloyd lab at stabilityai, 2023.
\newblock GitHub repository.

\bibitem[Labs(2023)]{qrmonster2023}
Monster Labs.
\newblock Controlnet qr code monster v2 for sd-1.5, 2023.

\bibitem[Liu et~al.(2021)Liu, Li, Du, Tenenbaum, and Torralba]{liu2021learning}
Nan Liu, Shuang Li, Yilun Du, Josh Tenenbaum, and Antonio Torralba.
\newblock Learning to compose visual relations.
\newblock \emph{Advances in Neural Information Processing Systems},
  34:\penalty0 23166--23178, 2021.

\bibitem[Liu et~al.(2022)Liu, Li, Du, Torralba, and
  Tenenbaum]{liu2022compositional}
Nan Liu, Shuang Li, Yilun Du, Antonio Torralba, and Joshua~B Tenenbaum.
\newblock Compositional visual generation with composable diffusion models.
\newblock In \emph{European Conference on Computer Vision}, pages 423--439.
  Springer, 2022.

\bibitem[Makowski et~al.(2021)Makowski, Lau, Pham, Paul~Boyce, and
  Annabel~Chen]{makowski2021parametric}
Dominique Makowski, Zen~J Lau, Tam Pham, W Paul~Boyce, and SH Annabel~Chen.
\newblock A parametric framework to generate visual illusions using python.
\newblock \emph{Perception}, 50\penalty0 (11):\penalty0 950--965, 2021.

\bibitem[Ngo et~al.(2023)Ngo, Sankaranarayanan, and Isola]{ngo2023clip}
Jerry Ngo, Swami Sankaranarayanan, and Phillip Isola.
\newblock Is clip fooled by optical illusions?
\newblock 2023.

\bibitem[Nichol et~al.(2021)Nichol, Dhariwal, Ramesh, Shyam, Mishkin, McGrew,
  Sutskever, and Chen]{nichol2021glide}
Alex Nichol, Prafulla Dhariwal, Aditya Ramesh, Pranav Shyam, Pamela Mishkin,
  Bob McGrew, Ilya Sutskever, and Mark Chen.
\newblock Glide: Towards photorealistic image generation and editing with
  text-guided diffusion models, 2021.

\bibitem[Oliva et~al.(2006)Oliva, Torralba, and Schyns]{oliva2006hybrid}
Aude Oliva, Antonio Torralba, and Philippe~G. Schyns.
\newblock Hybrid images.
\newblock \emph{ACM Trans. Graph.}, 25\penalty0 (3):\penalty0 527–532, 2006.

\bibitem[Owens et~al.(2014)Owens, Barnes, Flint, Singh, and
  Freeman]{owens2014camouflaging}
Andrew Owens, Connelly Barnes, Alex Flint, Hanumant Singh, and William Freeman.
\newblock Camouflaging an object from many viewpoints.
\newblock 2014.

\bibitem[Poole et~al.(2022)Poole, Jain, Barron, and
  Mildenhall]{poole2022dreamfusion}
Ben Poole, Ajay Jain, Jonathan~T. Barron, and Ben Mildenhall.
\newblock Dreamfusion: Text-to-3d using 2d diffusion.
\newblock \emph{arXiv}, 2022.

\bibitem[Radford et~al.(2021)Radford, Kim, Hallacy, Ramesh, Goh, Agarwal,
  Sastry, Askell, Mishkin, Clark, Krueger, and Sutskever]{radford2021clip}
Alec Radford, Jong~Wook Kim, Chris Hallacy, Aditya Ramesh, Gabriel Goh,
  Sandhini Agarwal, Girish Sastry, Amanda Askell, Pamela Mishkin, Jack Clark,
  Gretchen Krueger, and Ilya Sutskever.
\newblock Learning transferable visual models from natural language
  supervision.
\newblock In \emph{Proceedings of the 38th International Conference on Machine
  Learning}, pages 8748--8763. PMLR, 2021.

\bibitem[Ramesh et~al.(2021)Ramesh, Pavlov, Goh, Gray, Voss, Radford, Chen, and
  Sutskever]{ramesh2021zero}
Aditya Ramesh, Mikhail Pavlov, Gabriel Goh, Scott Gray, Chelsea Voss, Alec
  Radford, Mark Chen, and Ilya Sutskever.
\newblock Zero-shot text-to-image generation.
\newblock In \emph{International Conference on Machine Learning}, pages
  8821--8831. PMLR, 2021.

\bibitem[Rombach et~al.(2022)Rombach, Blattmann, Lorenz, Esser, and
  Ommer]{rombach2022ldm}
Robin Rombach, Andreas Blattmann, Dominik Lorenz, Patrick Esser, and Björn
  Ommer.
\newblock High-resolution image synthesis with latent diffusion models.
\newblock In \emph{Proceedings of the IEEE Conference on Computer Vision and
  Pattern Recognition (CVPR)}, 2022.

\bibitem[Saharia et~al.(2022)Saharia, Chan, Saxena, Li, Whang, Denton,
  Ghasemipour, Ayan, Mahdavi, Lopes, Salimans, Ho, Fleet, and
  Norouzi]{saharia2022imagen}
Chitwan Saharia, William Chan, Saurabh Saxena, Lala Li, Jay Whang, Emily
  Denton, Seyed Kamyar~Seyed Ghasemipour, Burcu~Karagol Ayan, S.~Sara Mahdavi,
  Rapha~Gontijo Lopes, Tim Salimans, Jonathan Ho, David~J Fleet, and Mohammad
  Norouzi.
\newblock Photorealistic text-to-image diffusion models with deep language
  understanding, 2022.

\bibitem[Shinbrot et~al.(2017)Shinbrot, Lazo, and Siu]{shinbrot2017network}
Troy Shinbrot, Miguel~Vivar Lazo, and Theo Siu.
\newblock Network simulations of optical illusions.
\newblock \emph{International Journal of Modern Physics C}, 28\penalty0
  (02):\penalty0 1750018, 2017.

\bibitem[Sohl-Dickstein et~al.(2015)Sohl-Dickstein, Weiss, Maheswaranathan, and
  Ganguli]{sohldickstein2015diffusion}
Jascha Sohl-Dickstein, Eric Weiss, Niru Maheswaranathan, and Surya Ganguli.
\newblock Deep unsupervised learning using nonequilibrium thermodynamics.
\newblock In \emph{Proceedings of the 32nd International Conference on Machine
  Learning}, pages 2256--2265, Lille, France, 2015. PMLR.

\bibitem[Song et~al.(2020)Song, Meng, and Ermon]{song2020denoising}
Jiaming Song, Chenlin Meng, and Stefano Ermon.
\newblock Denoising diffusion implicit models.
\newblock \emph{arXiv:2010.02502}, 2020.

\bibitem[Song et~al.(2021)Song, Sohl-Dickstein, Kingma, Kumar, Ermon, and
  Poole]{song2021scorebased}
Yang Song, Jascha Sohl-Dickstein, Diederik~P Kingma, Abhishek Kumar, Stefano
  Ermon, and Ben Poole.
\newblock Score-based generative modeling through stochastic differential
  equations.
\newblock In \emph{International Conference on Learning Representations}, 2021.

\bibitem[Tancik(2023)]{tancik2023illusions}
Matthew Tancik.
\newblock Illusion diffusion.
\newblock \url{https://github.com/tancik/Illusion-Diffusion}, 2023.

\bibitem[Ugleh(2023)]{ugleh2023spiral}
Ugleh.
\newblock Spiral town - different approach to qr monster.
\newblock
  \url{https://www.reddit.com/r/StableDiffusion/comments/16ew9fz/spiral_town_different_approach_to_qr_monster/},
  2023.

\bibitem[Wang et~al.(2023)Wang, Du, Li, Yeh, and Shakhnarovich]{wang2023score}
Haochen Wang, Xiaodan Du, Jiahao Li, Raymond~A Yeh, and Greg Shakhnarovich.
\newblock Score jacobian chaining: Lifting pretrained 2d diffusion models for
  3d generation.
\newblock In \emph{Proceedings of the IEEE/CVF Conference on Computer Vision
  and Pattern Recognition}, pages 12619--12629, 2023.

\bibitem[Wang et~al.(2020)Wang, Bylinskii, Hertzmann, and
  Pepperell]{wang2020toward}
Xi Wang, Zoya Bylinskii, Aaron Hertzmann, and Robert Pepperell.
\newblock Toward quantifying ambiguities in artistic images.
\newblock \emph{ACM Transactions on Applied Perception (TAP)}, 17\penalty0
  (4):\penalty0 1--10, 2020.

\bibitem[{Wikipedia contributors}()]{dress_wikipedia}
{Wikipedia contributors}.
\newblock The dress.
\newblock \url{https://en.wikipedia.org/wiki/The_dress}.
\newblock Accessed: November 9, 2023.

\bibitem[Zhang and Agrawala(2023)]{zhang2023controlnet}
Lvmin Zhang and Maneesh Agrawala.
\newblock Adding conditional control to text-to-image diffusion models, 2023.

\end{thebibliography}
}

\clearpage
\appendix

\section{Implementation Details}
\label{sec:apdx_impl}

We use the first two pixel-based stages of the DeepFloyd IF~\cite{deepfloyd2023} diffusion model. Specifically, we use the first stage which produces images of size $64\times 64$, and the second stage which upsamples images to $256\times 256$. Our method is applied in both stages, by implementing view transformations for both resolutions. DeepFloyd IF additionally predicts the variance, along with a noise estimate. We reduce multiple variance estimates by also taking a mean. We use a classifier guidance strength between 7 and 10, and between 30 and 100 inference steps depending on the prompt. We use the M size models for both stages.

Because DeepFloyd IF also estimates variances, we need to apply inverse views to these variance estimates, in addition to the noise estimates. For pixel permutation based views, we simply apply the inverse permutation to the variance estimates. For inversion, the inverse transformation would be negating the predicted logged variance, which does not make sense. We find that simply not inverting the variance estimates works well in this case.

DeepFloyd IF additionally uses a third super resolution stage, which is the Stable Diffusion x4 upscaler. This model upscales from $256\times 256$ to $1024\times 1024$. Because this model is a latent model, we do not apply our method to it. However, we find that we can use it with no modification to upscale our illusions without any loss in quality in the different views. We do this by upsampling conditioned on the prompt associated with the identity view. All results in \cref{fig:teaser} have been upsampled in this way.

\section{Dataset Collection}
\label{sec:apdx_dataset}

Our dataset consists of a list of styles, such as \texttt{"a street art of..."} or \texttt{"an oil painting of..."}, and a list of subjects such as \texttt{"an old man"} or \texttt{"a snowy mountain village"}. Subjects and styles were chosen by hand, using GPT-3.5 for inspiration. Prompt pairs are generated by randomly sampling a style prompt and prepending it to two randomly chosen subject prompts.

The CIFAR dataset was constructed by taking the 10 classes of CIFAR-10 as our subjects, and using the prompt \texttt{"a painting of"} as the style prompt. We take all 45 pairs of subjects, and prepend the style prompt to the subject prompts, resulting in 45 prompt pairs.

\section{Additional Results}
\label{sec:apdx_results}
\begin{figure}[t]
    \centering
    \includegraphics[width=\linewidth]{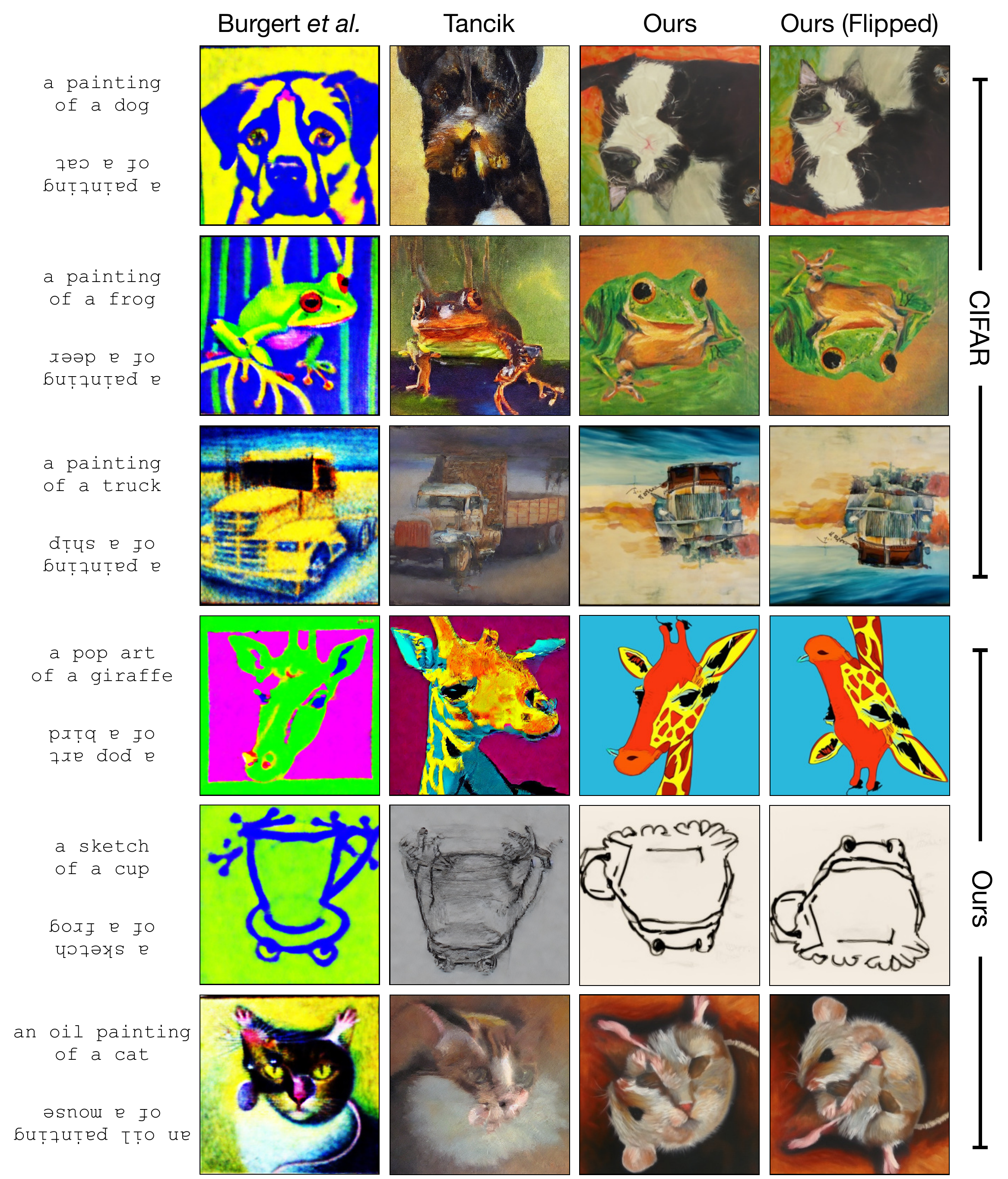}
    \caption{{\bf Qualitative Comparisons.} We compare more illusions generated by baselines to our illusions. We show examples from both our prompt dataset and the CIFAR prompt dataset.}
    \vspace{-1em}
\label{fig:comparison}
\end{figure}
\begin{figure}[t]
    \centering
    \includegraphics[width=\linewidth]{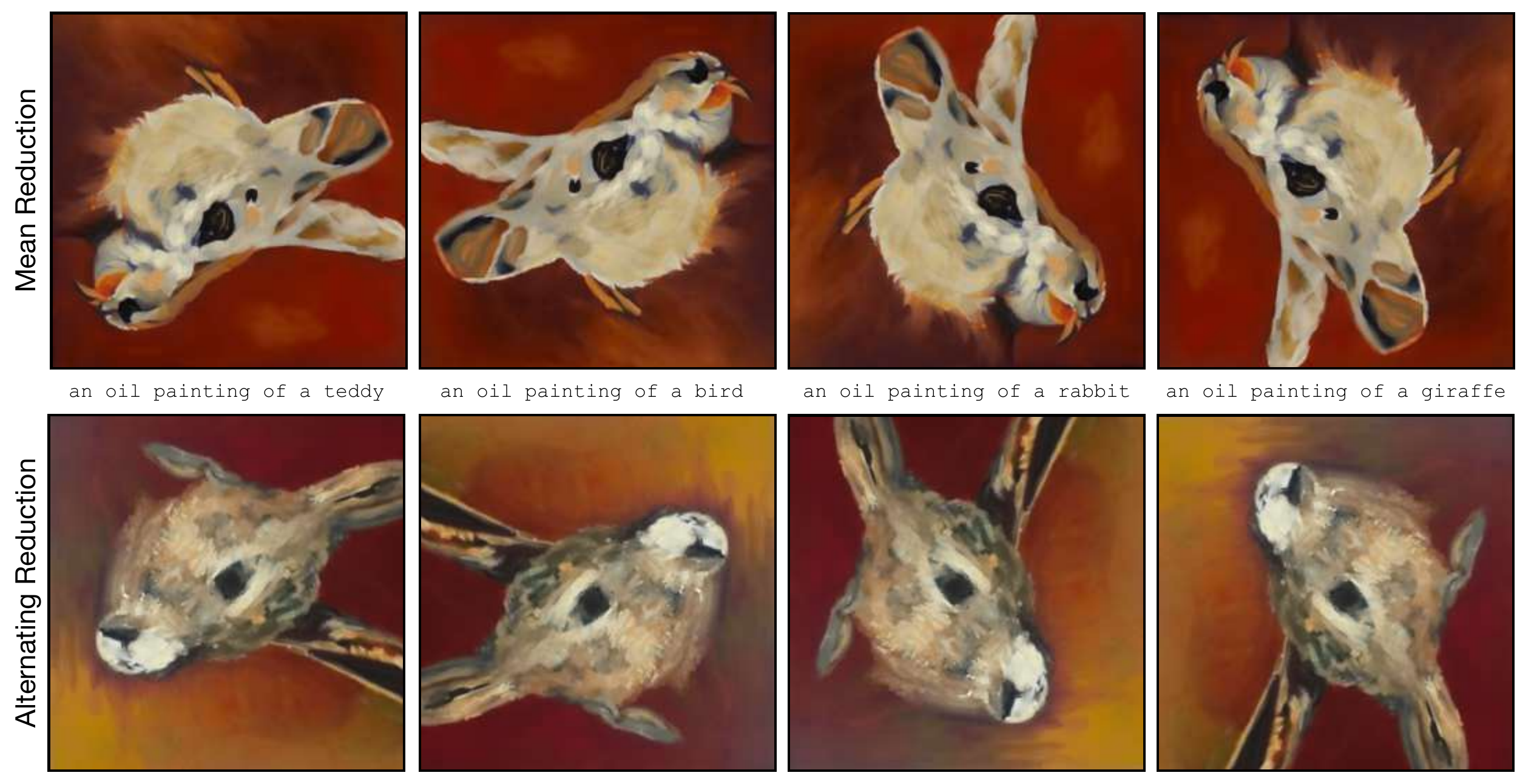}
    \caption{{\bf Combining Noise Estimates.} We show that mean reduction does better than alternating with an example of a 4-view sample image.}
    \vspace{-1em}
\label{fig:reduction}
\end{figure}
\begin{figure*}[t]
    \centering
    \includegraphics[width=\linewidth]{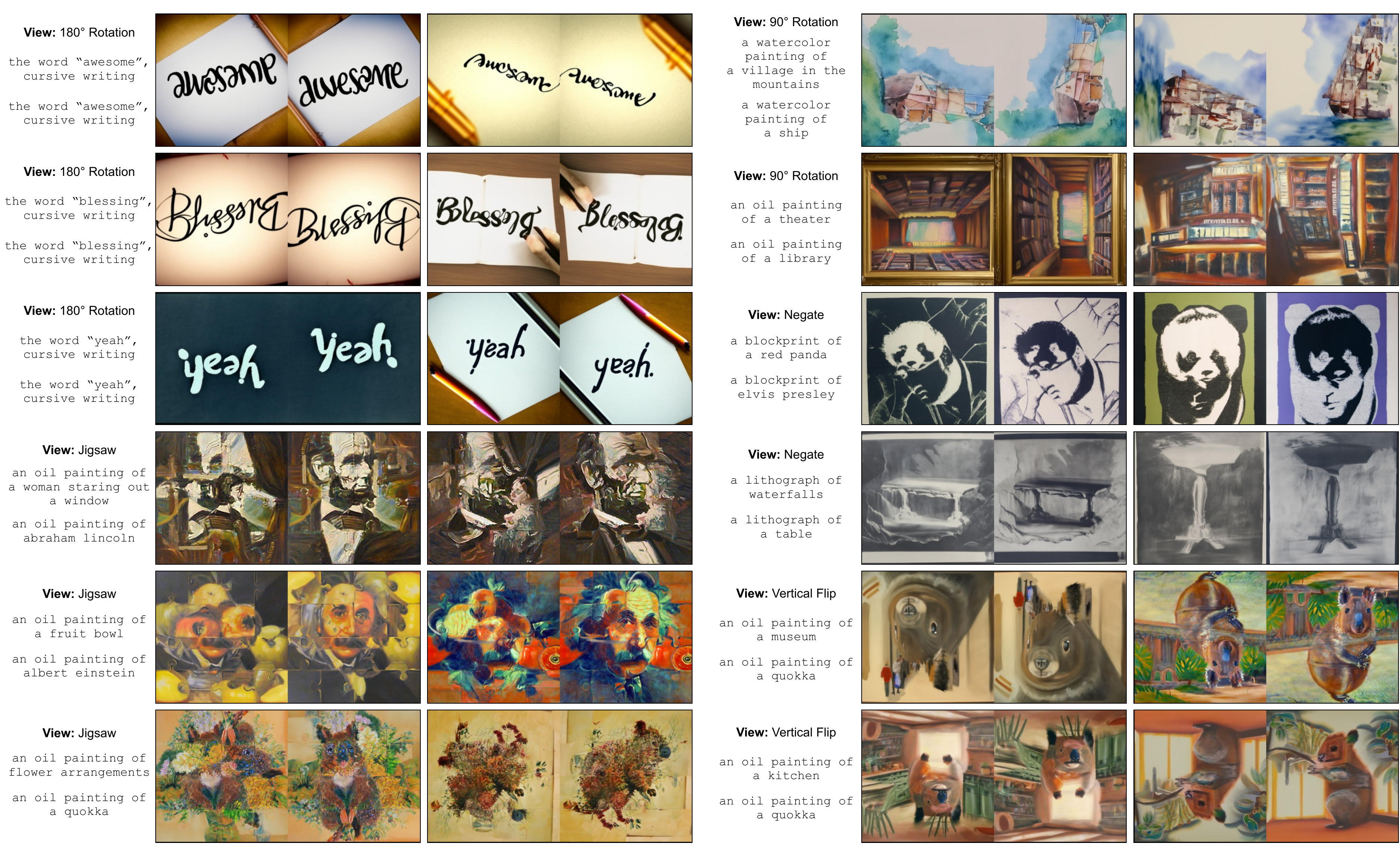}
    \caption{{\bf Qualitative Samples.} We show more illusions with views such as rotations, flips, color inversion, and jigsaw puzzles. }
    \vspace{-1em}
\label{fig:more_qualitative}
\end{figure*}

\begin{figure*}[t]
    \centering
    \includegraphics[width=\linewidth]{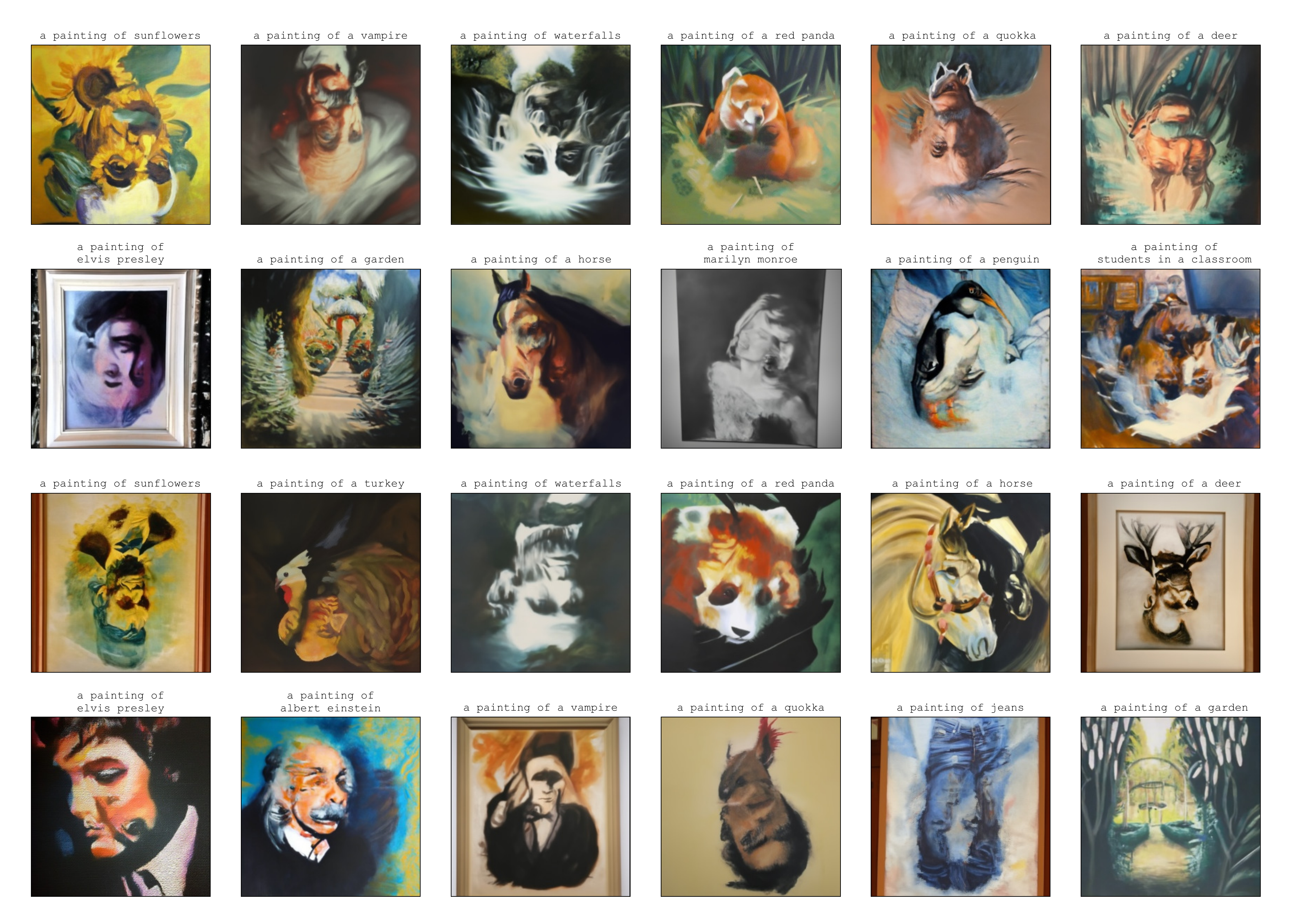}
    \caption{{\bf Flip illusions.} For each row, the prompt of the flipped image is the same. We encourage the reader to guess what the flipped prompt is. For an answer and flipped illusions, please see \cref{fig:flipped}.}
    \vspace{-1.5em}
\label{fig:einstein}
\end{figure*}
\begin{figure*}[t]
    \centering
    \includegraphics[width=\linewidth]{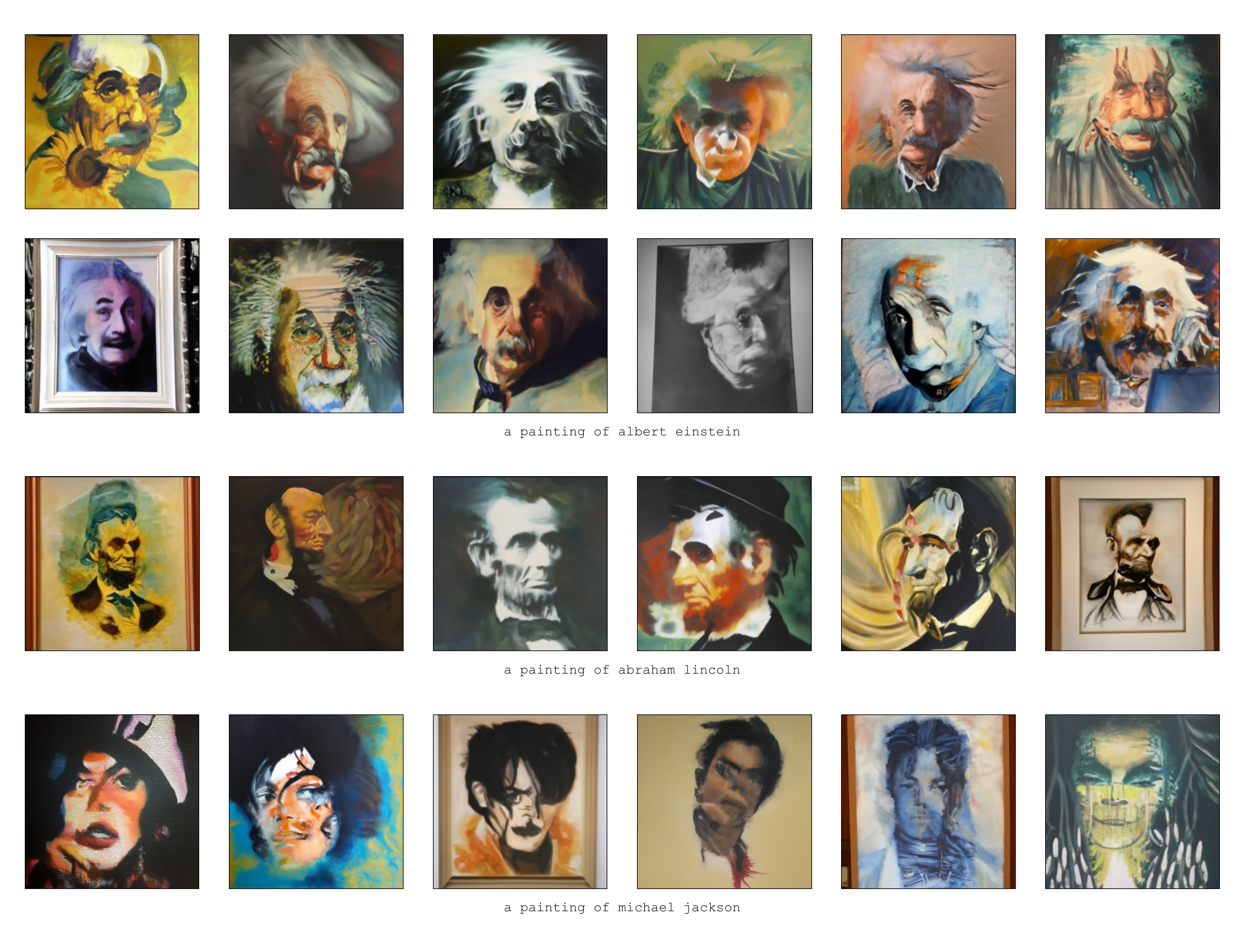}
    \caption{{\bf Flip illusions.} Flipped illusions from \cref{fig:einstein}, revealing the flipped prompt. Please refer to~\cref{fig:einstein} for the unflipped images.}
    \vspace{0.7em}
\label{fig:flipped}
\end{figure*}
We provide additional qualitative results in this section. In~\cref{fig:comparison}, we compare our method to baselines, using prompts from our dataset and the CIFAR prompt dataset. This is an extension of \cref{fig:qualitative}. We also generate more illusions with 90° and 180° rotations, ambigrams, ``polymorphic" jigsaw puzzles, color inversion, and vertical flips, which can all be found in~\cref{fig:more_qualitative}. In~\cref{fig:einstein}, we generate several flip illusions with the same flipped prompt, and different unflipped prompts, and we show flipped versions of these illusions in \cref{fig:flipped}.

\section{Random Samples}
\label{sec:apdx_random}
\begin{figure*}[t]
    \centering
    \includegraphics[width=\linewidth]{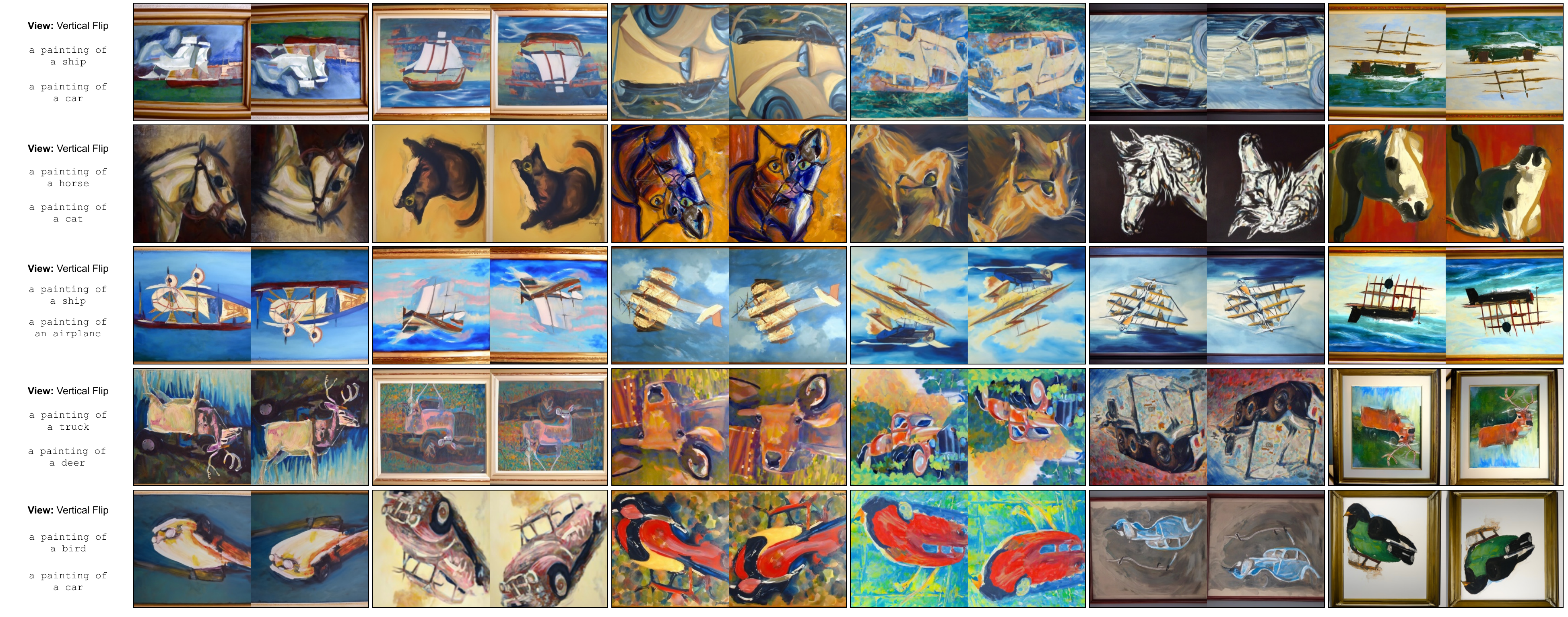}
    \caption{{\bf Random Samples.} We provide random samples for vertical flips using prompts from the CIFAR dataset. We show both views of the illusions side-by-side.}
\label{fig:random_cifar}
\end{figure*}
\begin{figure*}[t]
    \centering
    \includegraphics[width=\linewidth]{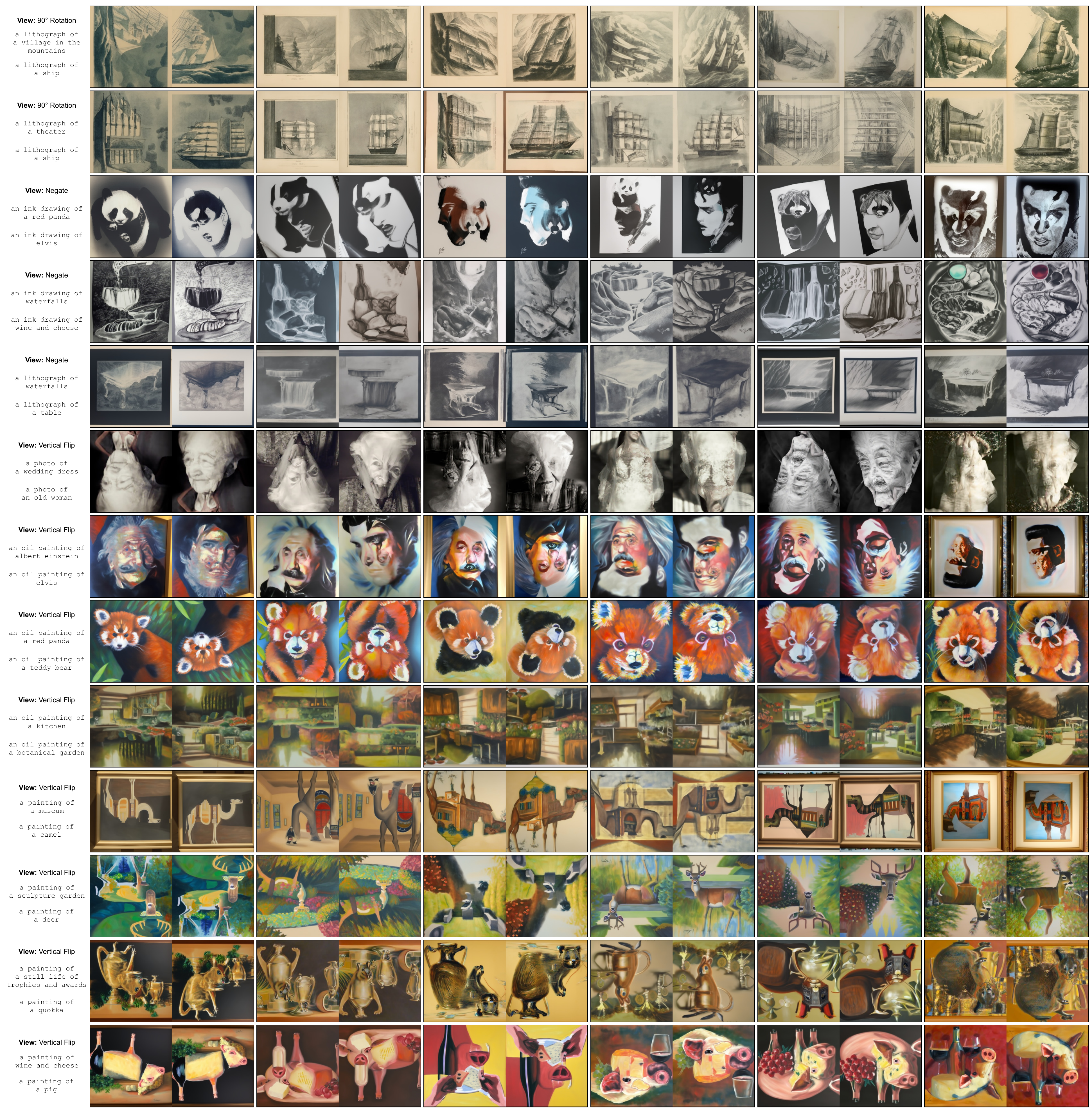}
    \caption{{\bf Random Samples.} We provide random samples for rotations, negations, and vertical flip views. We show both views of the illusions side-by-side.}
    \vspace{-1em}
\label{fig:random_apdx1}
\end{figure*}
\begin{figure*}[t]
    \centering
    \includegraphics[width=\linewidth]{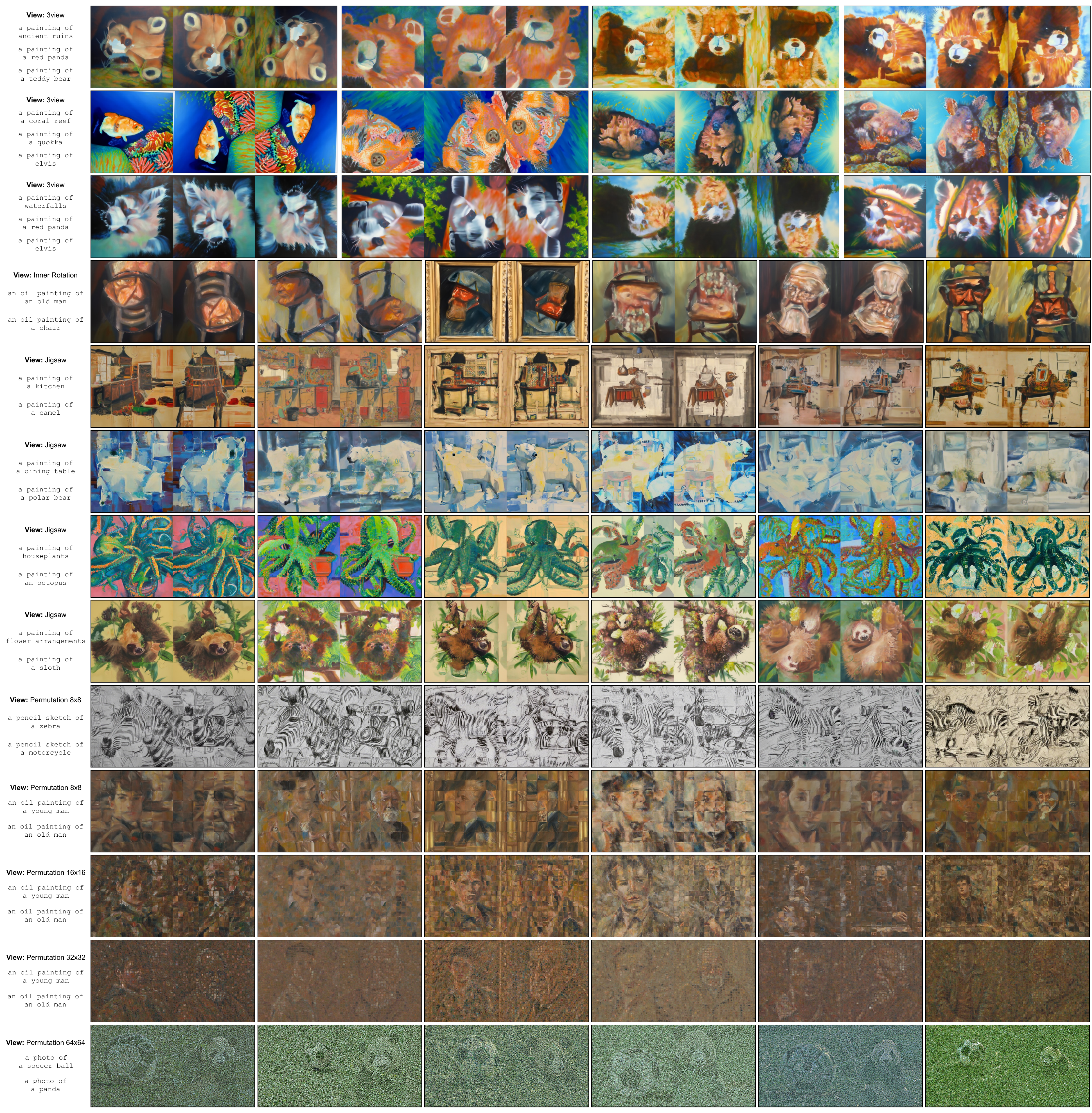}
    \caption{{\bf Random Samples.} We provide random samples for 3-view, inner rotation, jigsaw puzzle, and patch and pixel permutation views. We show all views of the illusions side-by-side.}
    \vspace{-1em}
\label{fig:random_apdx2}
\end{figure*}
We provide more random samples generated using our method. For rotations, color inversion, and vertical flips, please refer to~\cref{fig:random_apdx1}. For three-view, inner rotation, ``polymorphic" jigsaw puzzles, and patch and pixel permutation views, please refer to~\cref{fig:random_apdx2}.
We also provide random samples generated with prompts from the CIFAR dataset in~\cref{fig:random_cifar}.

The CIFAR prompt pair results, in~\cref{fig:random_cifar} as well as~\cref{tbl:quantitative}, are included as a proxy for random prompts. We note that systematically sampling truly random prompts for evaluation is tricky. Firstly, there is no standard method for sampling a random prompt. And secondly, not all prompt pairs make for good illusions. A straightforward example of this would be prompt pairs that differ in style. Therefore, evaluating illusion generation on completely random prompts may result in meaningless or misleading results, and as such prompts in~\cref{fig:random}, ~\cref{fig:random_apdx1}, and~\cref{fig:random_apdx2} are to some extent curated.

\section{The Art of Choosing Prompts}

We find that choosing good prompts is important to achieving good illusions. We lay out a few rules of thumb here. Firstly, it is very hard to reason as to what will make good illusions. Prompts that one may believe to work easily can fail consistently, and prompts that one may believe to have no chance of working may work fantastically. We find that more abstract styles, such as \texttt{"a painting"} or \texttt{"a drawing"} work much better than realistic styles such as \texttt{"a photo of"}. We believe this is because the constraints on realistic styles is too strong for illusions to work well. We also find that human faces make for good illusions, perhaps due to the sensitivity of the human visual system to face-like stimuli.

\section{Jigsaw Puzzle Implementation}
\label{sec:apdx_jigsaw}

We produce jigsaw puzzles by implementing a rearrangement of puzzle pieces as a permutation of pixels. We first hand-draw three puzzle pieces---a corner, edge, and center piece---such that they can disjointly tile a $64\times 64$, a $256\times 256$, or a $1024\times 1024$ image. All pieces in the puzzle are one of these three pieces, in different orientations. We then sample a random permutation of corner, edge, and center pieces respectively, and translate this permutation of pieces to a permutation of pixels.

\section{Combining Noise Estimates}
\label{sec:apdx_reduction}

Rather than taking the mean of noise estimates, we also experimented with alternating or cycling through noise estimates by timestep, as is done in~\cite{tancik2023illusions}. However, we find that this can lead to ``thrashing," in which the sample is optimized in different directions at different timesteps, leading to poor quality. Moreover, in illusions with more than two views, each view gets fewer denoising steps, resulting in lower quality illusions. For example, given four prompts each matched to a rotation of the image (i.e., \texttt{"a teddy"}, \texttt{"a bird"}, \texttt{"a rabbit"}, and \texttt{"a giraffe"}), the mean reduction outputs images with higher quality than the alternating method as shown in~\cref{fig:reduction}.

\section{Linearity of Views}
\label{sec:apdx_linearity}

As discussed in \cref{sec:views}, when a view $v$ is a linear transformation, it satisfies:
\begin{align}
    \label{eq:linear_cond_apdx}
    v(\x_t) &= v(w_t^{\text{signal}}\x_0 + w_t^{\text{noise}}\epsilon) \\
    &= w_t^{\text{signal}} v(\x_0) + w_t^{\text{noise}} v(\epsilon).
\end{align}
This is convenient because applying $v$ to the noisy image $\x_t$ is equivalent to applying $v$ to the signal, $\x_0$, and the noise, $\epsilon$, independently. In addition, the result is a linear combination of transformed signal and transformed noise, and is weighted as the diffusion model expects for timestep $t$.

However, there may be other conditions that work. For example, we could enforce
\begin{align}
    \label{eq:split_cond_apdx}
    v(\x_t) &= v(w_t^{\text{signal}}\x_0 + w_t^{\text{noise}}\epsilon) \\
    &= w_t^{\text{signal}} v_1(\x_0) + w_t^{\text{noise}} v_2(\epsilon),
\end{align}
with the interpretation being that $v$ somehow acts on the signal and noise in different ways, through $v_1$ and $v_2$, and combines them with the correct weightings. We leave this for future work.

\section{Statistical Consistency}
\label{sec:apdx_proof}

We provide a proof that for $\epsilon \sim \mathcal{N}(0, I)$ and square matrix $\A$, $\A\epsilon \sim \mathcal{N}(0, I)$ if and only if $\A$ is orthogonal, stated in~\cref{sec:views}. By properties of Gaussians, $\A\epsilon$ is also Gaussian, so we need only compute mean and covariances. The mean is given by
\newcommand{\Ex}{\mathbb{E}}
\begin{equation}
    \Ex[\A\epsilon] = \A\Ex[\epsilon] = 0.
\end{equation}
Because the mean is 0, the covariance is given by
\begin{align}
    \operatorname{Cov}(\A\epsilon) &= \Ex[(\A\epsilon)(\A\epsilon)^\intercal] \\
    &= \A\Ex[\epsilon\epsilon^\intercal]\A^\intercal \\
    &= \A\A^\intercal
\end{align}

So if $\A\epsilon \sim \mathcal{N}(0, I)$, then we must have $\operatorname{Cov}(\A\epsilon) = \A\A^\intercal = I$, or equivalently $\A$ must be orthogonal. And if $\A$ is orthogonal, then $\A\A^\intercal=I$ and $\A\epsilon \sim \mathcal{N}(0, I)$.

\vspace{5mm}

\begin{figure*}[t]
    \centering
    \includegraphics[width=0.95\linewidth]{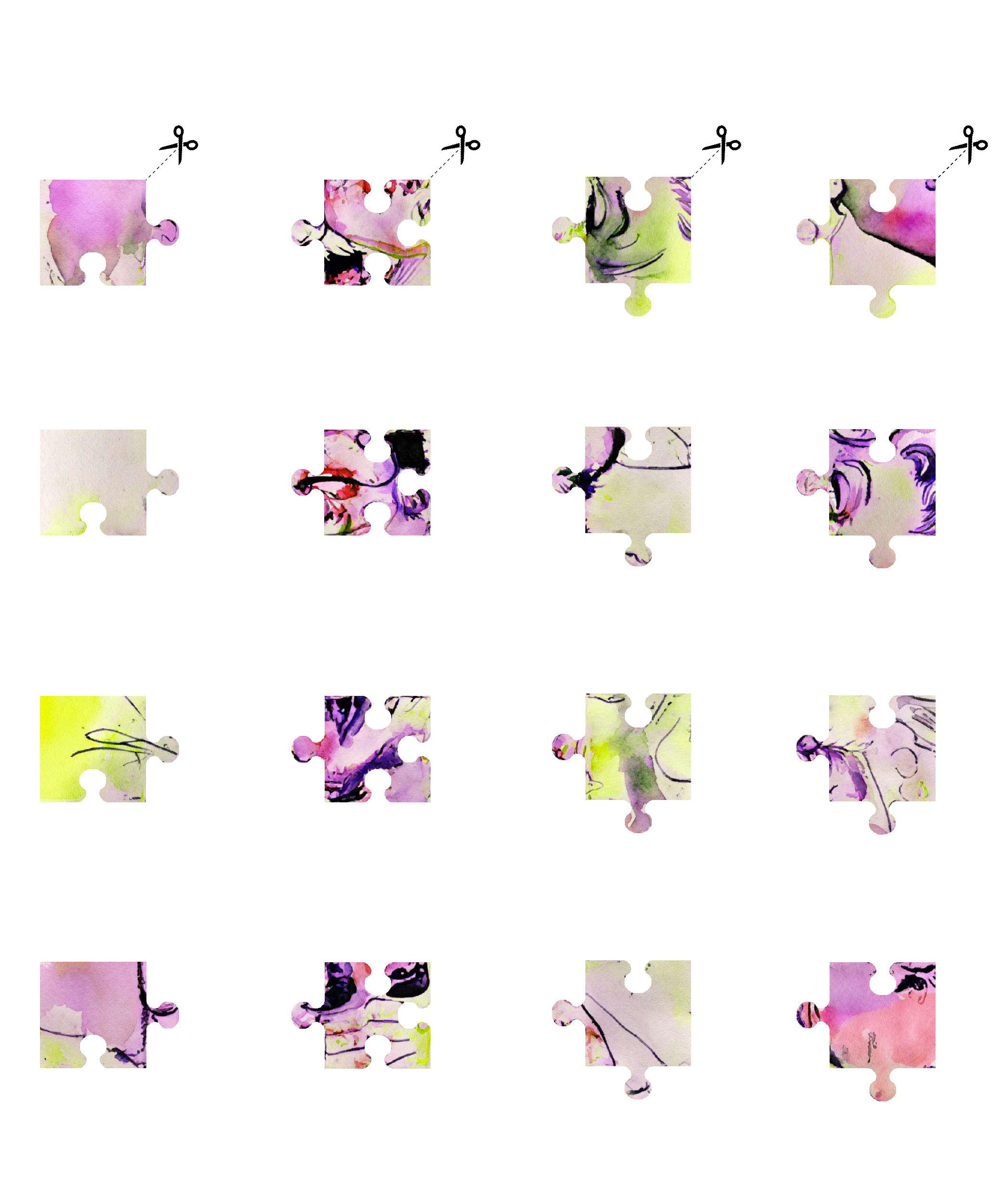}
    \caption{{\bf Cut Your Own Polymorphic Jigsaw!} We invite the reader to cut out their own polymorphic jigsaw puzzle, and try to discover both solutions.}
\label{fig:cutout}
\end{figure*}

\end{document}